\documentclass{article}

\PassOptionsToPackage{numbers,sort&compress}{natbib}
\usepackage[eandd, final]{neurips_2026}
\usepackage{amsmath}

\usepackage[utf8]{inputenc} %
\usepackage[T1]{fontenc}    %
\usepackage{hyperref}       %
\usepackage{url}            %
\usepackage{booktabs}       %
\usepackage{tabularx}       %
\usepackage{rotating}       %
\usepackage{amsfonts}       %
\usepackage{nicefrac}       %
\usepackage{microtype}      %
\usepackage{xcolor}         %
\usepackage{graphicx}
\graphicspath{{figures/}}
\usepackage{adjustbox}
\usepackage{enumitem}
\usepackage{placeins}       %
\title{Imperfectly Imperfect: Evaluating Graded Cognitive Diversity in LLM Persona Simulations}

\author{%
  Sion Weatherhead \\
  University of New South Wales, Sydney \\
  \texttt{s.weatherhead@unsw.edu.au} \\
  \And
  Ben Newell \\
  University of New South Wales, Sydney \\
  \texttt{ben.newell@unsw.edu.au} \\
  \AND
  Aaron Belbasis \\
  Aurecon Group, Melbourne \\
  \texttt{aaron.belbasis@aurecongroup.com.au} \\
  \And
  Flora Salim \\
  University of New South Wales, Sydney \\
  \texttt{flora.salim@unsw.edu.au} \\
}

\begin{document}

\maketitle

\begin{abstract}
Large language model (LLM) agents are increasingly used as synthetic participants in social, behavioural, and decision-making research, yet it remains unclear whether they can express graded cognitive diversity as assigned when engaging in simulated behaviour. We introduce a cognition-suite evaluation that tests whether simulated populations reproduce human variation in reasoning, self-reported traits, their associations, and downstream behaviour. Agents conditioned on synthetic or human-derived profiles of cognitive reflection, Need for Cognition, and Big Five personality answer the same suite as a cohort of 473 adults: two newly assembled tests, a Cognitive Reflection Test (CRT) and a Behavioural Reflection Test (BRT) whose decision scenarios measure reflection in behaviour, alongside an established CRT and the two trait questionnaires. Across seven full-battery simulation runs of four models, agents are able to recover Need for Cognition and Big Five trait distributions on self-report questionnaires, but fall short on simulating diversity of cognitive task performance and downstream behaviours. In the four Claude Sonnet 4.6 runs, every agent scores full marks on both CRTs, whatever reflection level its persona was assigned. In the other three runs agents make intuitive-lure errors (the tempting wrong answers CRT items invite), but 49 to 85\% score full marks, against under 10\% in the human sample. Claude Sonnet 5 comes closest to people's item-by-item pattern of lures; the other two runs give fewer lures and vary less between items. Configuration runs of the two CRTs on further models show that lure errors depend on the model and its thinking setting. In the BRT scenarios, every run over-produces 8 to 11 of 15 coded behaviours by at least 20 percentage points, and people's links between reflection and behaviour carry over only in part. Useful synthetic populations must preserve not only trait descriptors but the structured variation in reasoning, error type, and behaviour those descriptors are meant to organise.
\end{abstract}

\section{Introduction}
\label{sec:introduction}
Large language model (LLM) agents are increasingly used as synthetic participants in social, behavioural, and decision-making research, because they can produce language-mediated behaviour, maintain persona descriptions, and scale beyond human-data collection \citep{park2023, argyle2023, hewittPredictingResultsSocial, bisbee2024SyntheticReplacementsHuman, piao2025AgentSocietyLargeScaleSimulation}. Large-scale systems now simulate memory, planning, social interaction, economic activity, mobility, and policy-relevant social dynamics \citep{park2023, park2024GenerativeAgentSimulations, piao2025AgentSocietyLargeScaleSimulation}. Yet this progress leaves a basic validity problem unresolved: a synthetic population is not scientifically useful merely because agents sound plausible, match demographic marginals, or endorse personality-consistent self-descriptions.

For behavioural science, useful simulation requires preserving structured individual differences. Human populations vary in constructs such as cognitive reflection (the tendency to revise intuitive responses), enjoyment of effortful cognition, personality, and the downstream decisions and language these traits shape. Cognitive reflection is an especially sharp stress test because low-reflection behaviour is not simply lower accuracy. On Cognitive Reflection Tests (CRTs), humans predictably produce \textit{intuitive-lure errors}: compelling, fast, and \textit{specific} wrong responses that must be overridden to answer correctly \citep{frederick2005, thomson2016, pennycook2016}. Asked how many cubic feet of dirt are in a hole 3 feet deep, wide and long, many people intuitively calculate the answer ``27''; on reflection, they can realise a hole holds no dirt~\citep{thomson2016}. A model that answers correctly across all personas, refuses to answer incorrectly, or produces generic wrong answers can therefore fail the process being simulated even when its average score appears impressive. For synthetic populations, eliminating human-like error can be a fidelity failure.

Existing work has begun to test LLMs as social simulators, synthetic survey respondents, and psychologically shaped personas. LLMs can approximate some aggregate treatment-effect patterns in social-science experiments \citep{hewittPredictingResultsSocial}, but synthetic survey replacement remains fragile and can reproduce artefacts or exaggerated group differences \citep{bisbee2024SyntheticReplacementsHuman, wang2024Largelanguagemodels}. Persona-conditioning studies show that prompted Big Five traits can shift self-report (responses on psychometric surveys designed for the trait) and some linguistic markers \citep{jiang2024}, while psychometrically grounded lexical prompting and narrow psychometric fine-tuning suggest that trait-like behavioural profiles can sometimes be induced \citep{serapio-garcia2025psychometricframeworkevaluating, lulla2026DarkTriadModel}. These studies demonstrate steerability, but they do not yet establish whether LLM-based populations preserve the graded, joint cognitive structure assigned in simulation: marginal distributions, cross-construct associations, cognitive-error profiles, and downstream behavioural transfer. In people, such traits are also correlated with one another: those higher in Need for Cognition (NFC) tend to be more conscientious~\citep{weatherhead2026behavioural} and to score slightly higher on CRTs~\citep{pennycook2016,thomson2016}.  

We therefore evaluate LLM persona simulation as a problem of \emph{cognitive-diversity fidelity}. This follows construct-oriented approaches to AI evaluation, which argue that task scores or surface-level trait prompting are insufficient unless measured constructs predict behaviour coherently across contexts \citep{2026EvaluatingGeneralPurposeAIa, hagendorff2024MachinePsychology}. It also responds to evidence that contemporary instruction-tuned models may suppress human-like intuitive errors and reasoning biases \citep{hagendorff2023Humanlikeintuitivebehavior}. The important question is not whether agents appear believable, but whether a simulated population preserves the distributional, correlational, and behavioural structure of a human reference population.

We implement a cognition-suite evaluation that pairs established measures, CRT2, Need for Cognition and the ten-item Big Five Inventory (BFI-10), with two tests we built for it: a newly assembled bespoke CRT (bCRT) and Behavioural Reflection Test (BRT) decision scenarios, built to measure the behaviour associated with cognitive reflection at each assigned level. Need for Cognition measures enjoyment of effortful thinking \citep{cacioppo1984EfficientAssessmentNeed}; CRTs test reflective override \citep{frederick2005, thomson2016}; and prior work links cognitive reflection to misinformation sharing, moral judgement, risk preferences, open-minded thinking, and linguistic expression \citep{mosleh2021, pennycook2014roleanalyticthinking, pennycook2016, toplak2011CognitiveReflectionTest}. Linguistic Inquiry and Word Count (LIWC) diagnostics are used as supplementary language-based probes of BRT responses, motivated by prior work linking word categories to psychological expression in text \citep{tausczik2010PsychologicalMeaningWords, pennebaker,pennebaker2011secretlifepronouns}. In summary, these probes test whether agents can preserve cognitive diversity beyond questionnaire-style self-description.

The \textbf{contributions} of this paper are as follows:
\begin{enumerate}[noitemsep]
    \item We define \emph{cognitive-diversity fidelity} as an evaluation target for LLM persona simulation, covering marginal distributions, calibration, association structure, cognitive error-type fidelity, and downstream behavioural transfer.
    \item We introduce a cognition-suite evaluation that compares LLM persona simulations with a human study designed for this purpose; the bCRT and the BRT were developed and tested in the companion study~\citep{weatherhead2026behavioural}. No human responses to either were public before the knowledge cutoffs of the four models run on the full battery, so people's bCRT response distributions and coded BRT behaviours are fidelity targets absent from the models' training data.
    \item We compare human-reference data against seven LLM-agent populations from four models, and screen further models and thinking settings on the two CRTs in configuration runs.
    \item We show that current persona simulations recover questionnaire-style trait structure but fail stricter cognitive-diversity tests: on the two CRTs, scores concentrate at full marks, lure errors depend on the model and its thinking setting, and the two tests over-couple (their scores correlate far more strongly than in people); Claude Sonnet 5 comes closest to people's item-by-item lure pattern. Downstream BRT behaviour is compressed, and its link to reflection carries over only in part.
\end{enumerate}
\section{Related Work}
\label{sec:related_work}
\textbf{LLM agents and synthetic participants.}
LLM-based agents extend classical agent-based modelling by allowing simulated individuals to reason and communicate in natural language rather than through scalar states or hand-coded transition rules \citep{park2023, piao2025AgentSocietyLargeScaleSimulation}. In parallel, LLMs have been used as synthetic survey respondents and predictors of human experimental outcomes \citep{argyle2023, hewittPredictingResultsSocial}. However, aggregate outcome prediction is not the same as psychologically structured population simulation: a model may approximate mean responses while failing to preserve individual-difference distributions, covariance structure, or trait-linked behavioural variation \citep{crockett2023Shouldlargelanguage, wang2024Largelanguagemodels, aher2023}; \citet{hewittPredictingResultsSocial} predict condition-level effects from demographic profiles and \citet{park2024GenerativeAgentSimulations} replicate individuals from interviews, but neither specifies graded cognitive traits, nor tests whether a simulated population reproduces people's trait distributions and the traits' links to reasoning errors and decisions.
\textbf{Persona conditioning and psychometric shaping.}
Persona-conditioning studies~\citep{jiang2024, serapio-garcia2025psychometricframeworkevaluating, lulla2026DarkTriadModel, dash2025PersonaAssignedLargeLanguage} demonstrate steerability but typically do not test whether multiple cognitive and personality traits can be specified jointly, calibrated continuously, preserved as a distribution, and shown to transfer to downstream reasoning behaviour.

\textbf{Cognitive reflection as a simulation stress test.}
Cognitive reflection is unusually diagnostic because it requires more than plausible self-description. CRTs distinguish correct answers, intuitive-lure errors, and other failures \citep{frederick2005, thomson2016, pennycook2016}. This matters because higher cognitive reflection predicts real-world informational behaviour and linguistic patterns, including higher-quality information/news sharing and differences in cognitive, moral, political, and negative-emotion language \citep{mosleh2021}. It is also associated with moral judgement, risk preferences, open-minded thinking, and bias susceptibility \citep{pennycook2014roleanalyticthinking, toplak2011CognitiveReflectionTest}. For LLM simulation, this creates a direct test: can low-reflection personas produce human-like lure errors and downstream decision patterns, or do models collapse toward generic correctness? Prior evidence that ChatGPT-style systems suppress earlier LLMs' human-like intuitive biases makes this an especially likely failure mode \citep{hagendorff2023Humanlikeintuitivebehavior}.

\textbf{Evaluation as a construct-level problem.}
Our evaluation follows recent arguments that AI systems should be assessed through constructs that explain and predict behaviour across tasks, not only through isolated benchmark scores or self-report-like outputs \citep{2026EvaluatingGeneralPurposeAIa, hagendorff2024MachinePsychology}. It also aligns with robust-evaluation work showing that single prompt templates can misrepresent model behaviour and that distributions, quantiles, and psychometric tools can provide more stable evaluation targets \citep{polo2024Efficientmultipromptevaluation}. Because simulated agents should remain coherent across unfolding interactions, multi-turn consistency is also relevant; long-term dialogue evaluations show that LLMs still struggle with temporally extended consistency and memory \citep{maharana2024EvaluatingVeryLongTerm}. We therefore evaluate not one score, but a structured fidelity profile: marginals, calibration, association structure, intuitive-error categories, and behavioural transfer.
\section{Problem Statement: Cognitive-Diversity Fidelity}
\label{sec:problem_statement}
We evaluate each simulated population against the human reference sample using realised psychometric scores, coded BRT outcomes, and open-ended response features. The aim is to systematically test whether the simulated population preserves the structure present in humans.

We evaluate five fidelity targets. \emph{Marginal fidelity} tests whether realised agent distributions match human distributions for CRT2, bCRT, Need for Cognition, and Big Five traits, using Jensen--Shannon divergence, mean shifts, and variance ratios. \emph{Calibration fidelity} tests whether specified persona traits are realised in agent responses, using specified-to-realised slopes, $R^2$, RMSE, MAE, and signed error. \emph{Association fidelity} tests whether simulations preserve the human correlation structure among cognitive and personality constructs, summarised by simulated-minus-human Spearman correlation differences. \emph{Transfer fidelity} tests whether trait-outcome relationships observed in humans are reproduced in downstream BRT decisions and supplementary linguistic diagnostics. Finally, \emph{error-type fidelity} tests whether CRT responses preserve the human distribution over correct, intuitive-lure incorrect, other incorrect, and missing or unparseable responses.

\section{Evaluation Suite and Experimental Design}
\label{sec:methods}
\subsection{Human reference data}
\label{subsec:human_reference_data}
The human reference sample consisted of 473 Prolific adults (341 in the UK, 132 in the US) who completed an online Qualtrics study; the 471 who completed CRT2, Need for Cognition and the BFI-10 are the reference for those measures, and all 473 for bCRT scores and the BRT. The battery combined CRT2 \cite{thomson2016}, the four-item bCRT, the 18-item Need for Cognition scale \cite{cacioppo1984EfficientAssessmentNeed}, the BFI-10, demographics, and two open-ended BRT scenarios. CRT2 and bCRT answers were scored as correct, intuitive-lure incorrect, or other incorrect by item-specific answer keys, with a majority vote of three LLMs from different providers for answers the keys could not resolve (Appendix~\ref{app:crt_response_categories}). Many participants had seen the CRT2 items before, so the human CRT2 reference adjusts for familiarity: a regression on the number of CRT2 items each participant had seen gives every participant's predicted score, and mix of response categories, had they seen none (Appendix~\ref{app:human_reference_data}); raw CRT2 is reported beside it. bCRT is not adjusted: each bCRT item had been seen by 5--32\% of participants, against 41--75\% for each CRT2 item.

The BRT scenarios elicited written decisions and justifications in two evidence-conflict settings: a restaurant-booking scenario involving professional reviews, crowd reviews, and social obligation, and a medication scenario involving medical advice and conflicting social-media information. Two scenarios keep the battery short for participants, which protects response quality. Responses were coded by majority vote of three LLMs from different providers, with one fixed rubric for people and agents, into 15 coded behaviours (the BRT is presented in the companion study~\citep{weatherhead2026behavioural}), each labelled present or absent in a response, in four families: the sources a respondent plans to consult, the decision taken, the emotion expressed, and appeals to fairness or loyalty. The rubric, the definition of each behaviour and the panel's agreement are given in Appendix~\ref{app:brt_coding_rubric}.

As a supplementary diagnostic, each participant's two BRT responses were concatenated and scored with LIWC (Appendix~\ref{app:liwc_analysis}).
\subsection{Persona sources}
\label{subsec:persona_sources}
We evaluated two persona sources to separate direct profile instantiation from reusable synthetic-population simulation. \emph{Human-derived profiles} preserved the observed participant-level demographic and psychometric profiles, testing whether an LLM can realise real human trait profiles. \emph{Synthetic profiles} were sampled from a 5,000-profile bank estimated from the human reference data, testing whether fidelity survives an additional profile-generation layer rather than merely replaying observed rows. This distinction is important because LLMs are increasingly used as synthetic respondents or experimental proxies, but aggregate response prediction does not guarantee preservation of individual-difference structure \cite{argyle2023,hewittPredictingResultsSocial,bisbee2024SyntheticReplacementsHuman}. The synthetic bank used a Gaussian copula to preserve dependence structure among cognitive and personality variables, with targeted conditional sampling for demographic variables. Every synthetic run used the same 471 profiles, sampled without replacement to match the human reference sample size, and every human-derived run the 471 participants' own profiles. Each persona's prompt states its demographics, its five Big Five traits, its Need for Cognition and a cognitive-reflection level from 0 to 4, set at the 20th, 40th, 60th and 80th percentiles of people's familiarity-adjusted CRT2 scores; the bCRT is a newly assembled test with no human norms a model could have seen, so a bCRT level would carry no meaning; none is given, and the bCRT has no calibration target. Details of synthetic profile generation are provided in Appendix~\ref{app:persona_generation}.
\subsection{Agent simulation protocol}
\label{subsec:agent_simulation_protocol}
We ran seven simulation runs on four models (Appendix Table~\ref{tab:experimental-runs}). Claude Sonnet 4.6 (no extended thinking) was run in four conditions crossing persona source with prompt format, to test whether how traits are stated interacts with where the profile comes from: \emph{numeric} prompts give each trait as a scale value, \emph{lexical} prompts as a graded verbal descriptor (see Appendix~\ref{app:prompt_screening} for the full rationale). Claude Sonnet 5 (adaptive thinking), Gemini 3.5 Flash (thinking level minimal) and DeepSeek V4-Pro (reasoning effort high, via Together AI), three models from three providers that produce intuitive-lure errors in the configuration runs, were each run once with synthetic profiles in numeric format, chosen for Gemini 3.5 Flash and DeepSeek V4-Pro by a criterion set before the runs (Appendix~\ref{app:prompt_screening}). Each persona completed the same task battery as the human reference sample. The administration used one model call per block, with block order:
\[
\mathrm{BRT} \rightarrow \mathrm{Reasoning} \rightarrow \mathrm{Need\ for\ Cognition} \rightarrow \mathrm{BFI}.
\]
The reasoning block contained CRT2, bCRT, and decoy questions in randomised order. Conversation history was retained across blocks for each persona, preserving a modest form of temporal continuity while keeping the evaluation scoreable at the block level. To check whether the BRT answers, written first, shape the reasoning answers through the retained history, we gave the reasoning block again to 150 profiles with no prior conversation: Claude Sonnet 4.6 produced no lures with or without history (Appendix~\ref{app:prompt_screening}). Configuration runs compare models, thinking settings and prompt formats on the eight CRT2 and bCRT items, on one fixed panel of 150 synthetic profiles (Appendix~\ref{app:prompt_screening}).
\subsection{Statistical analyses}
\label{subsec:statistical_analyses}
We compared each agent condition against the human reference sample using five fidelity families. This multi-metric design follows construct-oriented AI evaluation, where a construct is informative only if it predicts coherent behaviour across tasks, and robust LLM evaluation work showing that single-score summaries can hide instability \cite{2026EvaluatingGeneralPurposeAIa,polo2024Efficientmultipromptevaluation}.

Each family is measured as Section~\ref{sec:problem_statement} defines it. Rates carry Wilson 95\% intervals, except lure rates pooled over a test's four items, which carry persona-clustered bootstrap intervals (5,000 resamples); odds ratios carry Wald intervals (profile-likelihood intervals for Firth-penalised models where events are few), and mean shifts Welch intervals.

Behavioural fidelity was evaluated on the 15 coded behaviours by the prevalence gap between run and humans, with 2,000-resample bootstrap intervals on the gap (\emph{compressed}: a gap of at least 20 points with non-overlapping 95\% intervals), by \emph{steering}, the change in each behaviour with the reflection level the persona was told (an odds ratio per level), and by transfer of the human bCRT association, read from the interval on the difference between run and human slopes; where people show no detectable association and the run's own interval excludes 0, a failure to transfer means the run \emph{introduces} one, as for trait correlations; models with fewer than 20 events or non-events were flagged \emph{unstable}. Trait-structure transfer applies the same interval logic to the 28 construct-pair correlations, with 10,000-resample intervals on $\rho$ and $\Delta\rho = \rho_{\mathrm{run}} - \rho_{\mathrm{human}}$; the larger count keeps the interval ends stable where they decide a verdict. Each comparison of a run with people then receives a verdict, a label for whether the run reproduces people's value or how it departs from it (for example \emph{preserved}, \emph{weakened}, \emph{reversed} or \emph{introduced}); Appendix~\ref{app:formal_fidelity_definitions} defines each verdict and the order in which they are assigned. \emph{Preserved} means not detectably different, not shown identical. Verdicts are read from intervals without correction across pairs or outcomes, so about one target in twenty is expected to read as a departure even under perfect fidelity. Supplementary LIWC-transfer models are reported in Appendix~\ref{app:liwc_relationship_fidelity}.
\section{Results}
\label{sec:results}
\subsection{Self-reported traits are recoverable, but reflection measured by performance is not}
\label{sec:results_distribution_fidelity}
\begin{figure}[t]
  \centering
  \includegraphics[width=\linewidth]{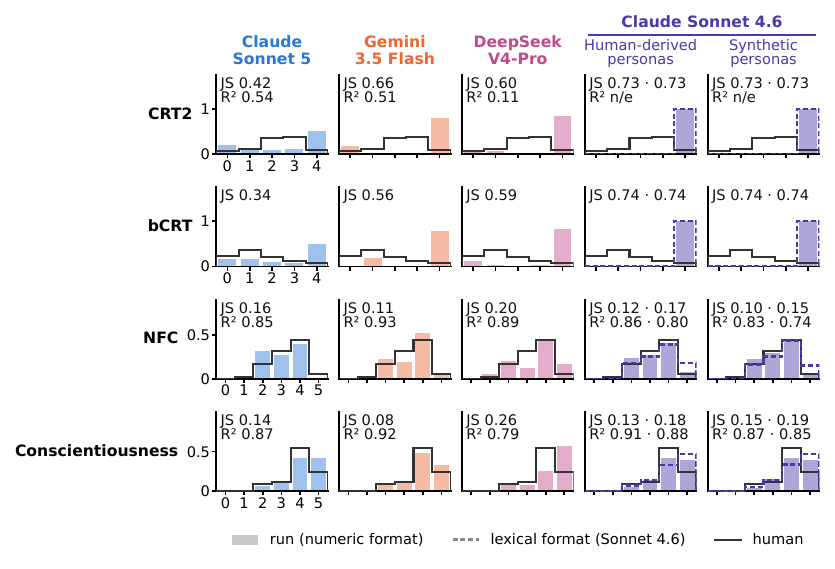}
  \caption{Do the runs reproduce the human score distributions of the four focal constructs? Each panel sets one run against the human distribution. Corners give the Jensen--Shannon divergence from the human distribution (JS) and the calibration $R^2$ of realised on specified score (bCRT was never specified); in the Claude Sonnet 4.6 panels the numeric value comes first. n/e: not estimable, because every agent scored 4/4 and there is no variation to calibrate. The other four Big Five traits are in Appendix Figure~\ref{fig:distribution_grid}. $N = 471$ per run; human $N = 471$ (bCRT 473).}
  \label{fig:distribution_fidelity}
\end{figure}

Figure~\ref{fig:distribution_fidelity} shows a sharp split between questionnaire-style traits and the two CRTs. Marginal fidelity is strongest for NFC and the Big Five traits in every run: JS divergence from the human distribution is $.10$--$.20$ for NFC and $.11$--$.24$ averaged over the Big Five, against $.34$--$.74$ for CRT2 and bCRT (Appendix Figure~\ref{fig:distribution_grid}). Under numeric prompting, Claude Sonnet 4.6 shifts the NFC mean by only $+0.01$ points with either persona source. The major distortions are shifts in dispersion and category mass, not wholesale loss of the scale.
Specified-to-realised calibration shows the same split (Appendix Figure~\ref{fig:calibration_fingerprint}). Across the seven runs, calibration $R^2$ is $.74$--$.93$ for NFC and $.79$--$.96$ for the Big Five traits.

By contrast, CRT2 and bCRT depart from people in every run, in two different ways. In the four Claude Sonnet 4.6 runs every agent scores 4 of 4 on both tests, whatever reflection level its persona specifies: scores shift upward by 1.53 (CRT2) and 2.56 (bCRT) points on a four-item scale, variance is zero, and calibration $R^2$ cannot be estimated. The other three runs vary on both tests and partly track the specified level, but their scores concentrate at full marks: on bCRT, 49--82\% of their agents score 4 of 4 and 7--35\% score 1 to 3, against 8\% and 69\% of people, with the same pattern on CRT2 (Figure~\ref{fig:distribution_fidelity}). Outside full marks, most of their agents score 0 or 1, so their variance exceeds people's (CRT2 variance ratio $1.21$--$2.78$). Claude Sonnet 5's CRT2 mean does not differ detectably from people's (shift $+0.16$, 95\% interval $-0.01$ to $+0.33$), yet its distribution still diverges.
\begin{table}[t]
\centering
\footnotesize
\setlength{\tabcolsep}{4pt}
\caption{Fidelity scorecard: run-level summaries by condition.}
\label{tab:scorecard}
\begin{tabularx}{\linewidth}{@{}>{\hsize=1.00\hsize\raggedright\arraybackslash}Xrrrrrrrc@{}}
\toprule
Condition & \begin{tabular}[b]{@{}c@{}}CRT2 \\ gap\end{tabular} & \begin{tabular}[b]{@{}c@{}}bCRT \\ gap\end{tabular} & \begin{tabular}[b]{@{}c@{}}Calib. \\ RMSE\end{tabular} & \begin{tabular}[b]{@{}c@{}}Assoc. \\ MAE\end{tabular} & \begin{tabular}[b]{@{}c@{}}BRT \\ MAE\end{tabular} & \begin{tabular}[b]{@{}c@{}}Com- \\ pressed\end{tabular} & \begin{tabular}[b]{@{}c@{}}Un- \\ stable\end{tabular} & \begin{tabular}[b]{@{}c@{}}Transfer / \\ fail / indet.\end{tabular} \\
\midrule
Claude Sonnet 5 & $-$0.001 & $-$0.156 & 0.458 & 0.085 & 0.263 & 8 & 5 & 3 / 3 / 4 \\
Gemini 3.5 Flash & $-$0.150 & $-$0.253 & 0.445 & 0.088 & 0.244 & 8 & 5 & 3 / 4 / 3 \\
DeepSeek V4-Pro & $-$0.230 & $-$0.280 & 0.690 & 0.101 & 0.369 & 11 & 4 & 1 / 3 / 7 \\
Claude Sonnet 4.6 · human · numeric & $-$0.333 & $-$0.411 & 0.534 & 0.077 & 0.309 & 9 & 6 & n/e \\
Claude Sonnet 4.6 · human · lexical & $-$0.333 & $-$0.411 & 0.612 & 0.072 & 0.324 & 10 & 6 & n/e \\
Claude Sonnet 4.6 · synthetic · numeric & $-$0.333 & $-$0.411 & 0.555 & 0.084 & 0.319 & 10 & 7 & n/e \\
Claude Sonnet 4.6 · synthetic · lexical & $-$0.333 & $-$0.411 & 0.629 & 0.096 & 0.327 & 10 & 6 & n/e \\
\bottomrule
\end{tabularx}
\par\smallskip
\begin{minipage}{\linewidth}\footnotesize\raggedright
Gap = run minus human intuitive-lure prevalence (negative: fewer lures than people; human reference: the familiarity-adjusted rate for CRT2 and the rate as answered for bCRT). Calib.\ RMSE = mean over the seven specified constructs. Assoc.\ MAE = mean absolute difference in pairwise correlations over CRT2, bCRT, NFC, Conscientiousness, Openness and Neuroticism (15 pairs; 6 for the Claude Sonnet 4.6 conditions, whose CRT2 and bCRT pairs are undefined). BRT MAE = mean absolute prevalence gap over the 15 coded BRT behaviours; compressed, unstable and the transfer verdicts (transfers / fails / indeterminate) count those behaviours as defined in Section~\ref{subsec:statistical_analyses}; n/e: not estimable. DeepSeek V4-Pro BRT N = 467: four personas returned no usable open-ended response, and the provider has since retired the model snapshot.\par
\end{minipage}
\end{table}

\subsection{Prompt format changes questionnaire fidelity, not reflection}
\label{sec:results_prompt_tradeoffs}
In the four Claude Sonnet 4.6 conditions, numeric prompts gave closer questionnaire fidelity than lexical prompts: lower NFC divergence (JS $.10$ and $.12$ vs.\ $.15$ and $.17$), lower average BFI divergence ($.14$ vs.\ $.17$ and $.18$) and smaller calibration error (Table~\ref{tab:scorecard}). The advantage does not reach reflection or behaviour: under both formats every agent scores 4 of 4 on CRT2 and bCRT (Section~\ref{sec:results_distribution_fidelity}), and both lexical conditions have a higher BRT prevalence error than both numeric conditions (Table~\ref{tab:scorecard}).

\subsection{Intuitive-lure errors depend on model and thinking regime}
\label{sec:results_intuitive_lures}
In people, CRT errors are structured: on bCRT, $41.1\%$ of answers are intuitive lures and $22.8\%$ other errors (Appendix Table~\ref{tab:human-crt2-reference} gives the full human response mix).

\begin{figure}[t]
  \centering
  \includegraphics[width=\linewidth]{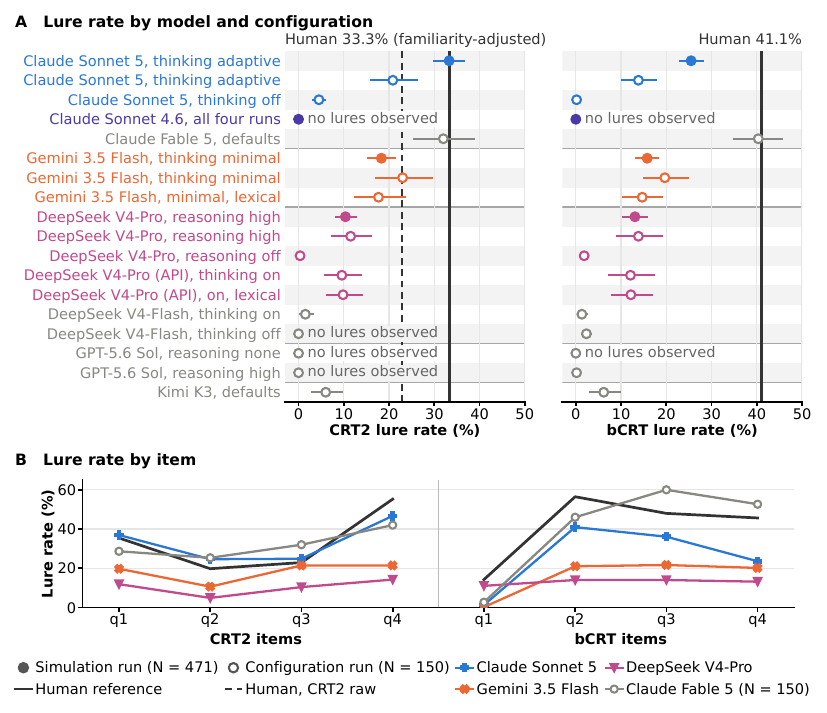}
  \caption{Which runs produce intuitive-lure errors, and on which items? (A) Lure rate on CRT2 and bCRT with its 95\% interval, for the simulation runs ($N = 471$) and for the configuration runs (Section~\ref{subsec:agent_simulation_protocol}). (B) Lure rate per item for people (CRT2 familiarity-adjusted) and the three runs that produce lures, with Claude Fable 5, the configuration run whose lure rates are closest to people's. Models with configuration runs only are drawn in grey. Intervals are persona-clustered bootstrap intervals. Solid black lines are the human reference for each test (CRT2: familiarity-adjusted, people's rate predicted for items not seen before; bCRT: as answered); the dashed line is people's raw CRT2 rate.}
  \label{fig:crt_bcrt_intuitive_lure_fidelity}
\end{figure}
Agents reproduce this structure only in part (Figure~\ref{fig:crt_bcrt_intuitive_lure_fidelity}). The four Claude Sonnet 4.6 runs produce no lures: every CRT2 and bCRT response is correct. The other three runs do. Claude Sonnet 5's pooled CRT2 lure rate is $33.3\%$, as is people's familiarity-adjusted rate, and on both tests it puts the same items highest and lowest as people do, but half its agents score full marks, so its lures come from the other half (Section~\ref{sec:results_distribution_fidelity}); on bCRT it gives fewer lures than people on every item ($25.5\%$ against $41.1\%$ overall). Gemini 3.5 Flash and DeepSeek V4-Pro produce fewer lures than people's bCRT and familiarity-adjusted CRT2 rates, with less difference between items than people show. On the configuration ladder (Appendix Table~\ref{tab:configuration-ladder}), turning thinking off on the same 150 profiles removes most lures in Claude Sonnet 5 and DeepSeek V4-Pro, and GPT-5.6 Sol produces one lure in 2,400 responses across its two reasoning efforts.
\subsection{Most trait correlations survive, but the two reflection tests over-couple}
\label{sec:results_trait_structure}
\begin{figure}[t]
  \centering
  \includegraphics[width=\linewidth]{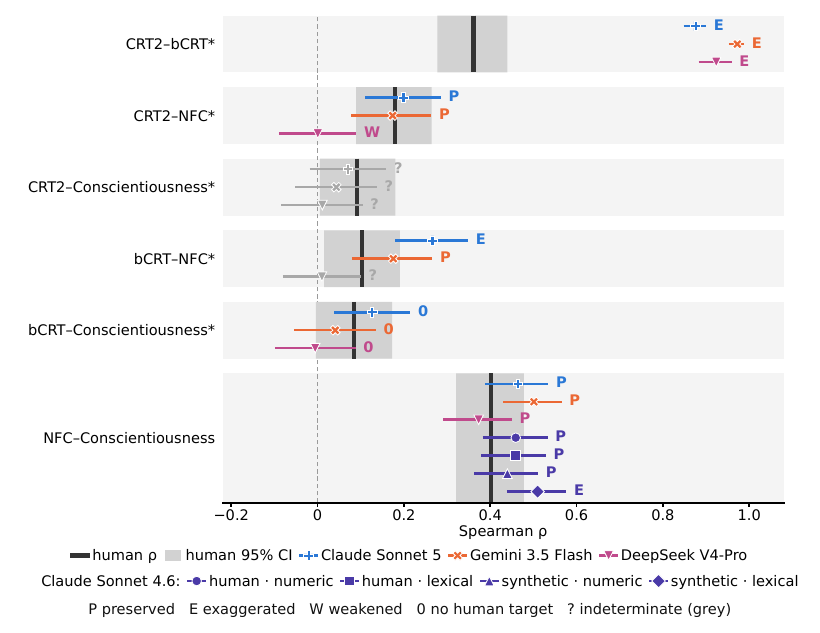}
  \caption{Does each run reproduce the human correlations among the four focal constructs? Each band is one pair of constructs: the black bar and grey band give the human Spearman correlation and its 95\% interval; each marker gives one run's correlation with its 95\% interval and its verdict (rules in Appendix~\ref{app:formal_fidelity_definitions}). ``0'': people show no detectable correlation, so there is nothing to reproduce; ``?'': people do, but the run's interval is too wide to tell whether it kept the correlation or lost it. *The four Claude Sonnet 4.6 conditions are not estimable on these pairs, because every agent scored 4/4 on CRT2 and bCRT. $N = 471$ per side; 95\% percentile bootstrap intervals from 10,000 resamples. Appendix Figure~\ref{fig:f4_all_pairs} gives the differences from the human values for all 28 pairs.}
  \label{fig:association_fidelity}
\end{figure}

Most construct pairs keep their human correlation (Figure~\ref{fig:association_fidelity}). Of the 16 pairs with a human target, Claude Sonnet 5 preserved 11, departed from 3 (exaggerated or weakened) and left 2 indeterminate; Gemini 3.5 Flash 13, 1 and 2; DeepSeek V4-Pro 10, 3 and 3. Of the 12 such pairs estimable in the Claude Sonnet 4.6 conditions, the two human-derived conditions each preserved 11 and departed from 1, the two synthetic conditions 9, 2 and 1; the 13 pairs involving CRT2 or bCRT are not estimable there. No run reversed a human correlation, and Claude Sonnet 5 and both synthetic Claude Sonnet 4.6 conditions introduced one that people do not show (NFC--Agreeableness).

The largest departure is the coupling of the two reflection tests: CRT2 and bCRT correlate at .36 in people but at .88--.97 in the three runs that vary on them. Outside the focal pairs, all four Claude Sonnet 4.6 conditions exaggerated NFC--Openness (human .29; $\Delta\rho$ from $+0.18$ to $+0.29$), and Claude Sonnet 5, DeepSeek V4-Pro and Claude Sonnet 4.6 synthetic-numeric weakened Conscientiousness--Extraversion (human .18) to about zero (Appendix~\ref{app:trait_structure_transfer}).
\subsection{BRT behaviour is compressed, and its link to reflection transfers only in part}
\label{sec:results_brt_compression}
\begin{figure}[t]
  \centering
  \includegraphics[width=\linewidth]{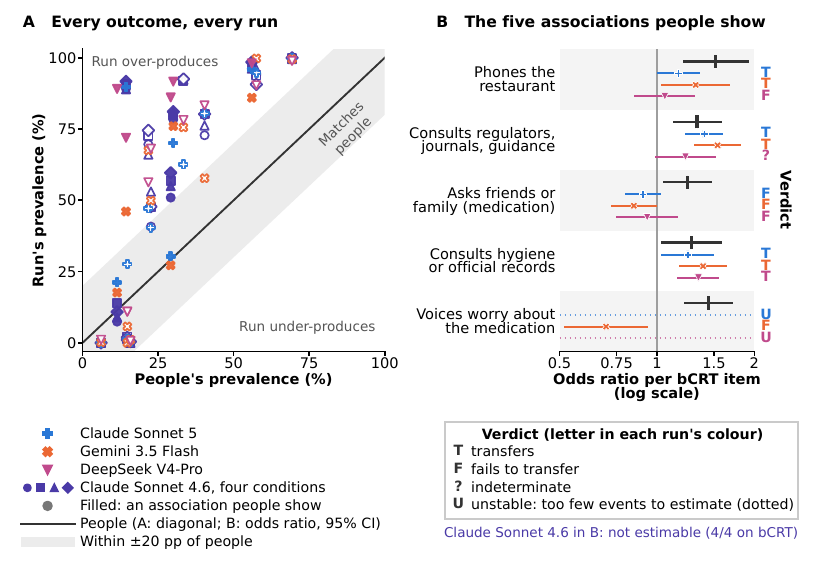}
  \caption{Does human behavioural variation survive, and does its link to reflection carry over? (A) Prevalence of each of the 15 coded BRT behaviours in each run against its prevalence in people; the diagonal marks equal prevalence and the grey band $\pm20$ percentage points. Filled markers are the five behaviours for which people show a reliable association with bCRT (Benjamini--Hochberg $q < .05$ within family). (B) For those five, the odds ratio per bCRT item in people and in each run whose bCRT scores vary, with 95\% intervals. The verdict tests the difference between run and human slopes (Section~\ref{subsec:statistical_analyses}), so overlapping intervals can still differ. Appendix Figure~\ref{fig:brt_prevalence_by_run} gives panel A run by run. $N = 471$ per run (DeepSeek V4-Pro 467); people $N = 473$.}
  \label{fig:brt_behavioural_compression}
\end{figure}

The BRT tests whether persona conditioning carries through to downstream behaviour, beyond psychometric test scores. Every run compresses the human behavioural profile (Figure~\ref{fig:brt_behavioural_compression}A): of the 15 coded behaviours (Section~\ref{subsec:human_reference_data}), 8 to 11 per run statistically differ from people by at least 20 percentage points (Table~\ref{tab:scorecard}), and every such difference is an excess. On the nine behaviours people show at least $20\%$ of the time, where a shortfall of that size was possible, no run falls more than 2 points below people.

Agents consult authoritative sources and make moral appeals far more often than people. In the restaurant scenario, agents keep the booking more often and pair this with appeals to fairness or loyalty far more often than people, and with rebuke above the human rate in every run. In the medication scenario, agents consult regulators and official guidance, and voice worry, far more often than people.

Whether people's link between reflection and behaviour carries over is more mixed (Figure~\ref{fig:brt_behavioural_compression}B). Of the five associations people show reliably, Claude Sonnet 5 and Gemini 3.5 Flash reproduce three and DeepSeek V4-Pro one; all three runs fail on asking friends or family about the medication (Appendix Table~\ref{tab:brt-transfer} gives the verdicts on all 15 behaviours). Transfer cannot be tested in the Claude Sonnet 4.6 runs, whose bCRT scores do not vary. Where people show no detectable association, the runs can introduce one: all three link reflection to verifying the video's cited sources. Steering (Section~\ref{subsec:statistical_analyses}) can be estimated in every run, Claude Sonnet 4.6 included (Appendix Table~\ref{tab:brt-steering} and Figure~\ref{fig:brt_level_prevalence}): personas told a higher level consult regulators more often in all seven runs, check hygiene or official records more often in six and phone the restaurant more often in two, as more reflective people do, but ask friends or family about the medication less often in five, where more reflective people ask more often. Worry about the medication is the exception: six runs voice it in 96--99\% of responses, so its steering model is unstable (Section~\ref{subsec:statistical_analyses}), though three Claude Sonnet 4.6 conditions drop to 87--93\% at the highest told level. In Gemini 3.5 Flash, the one stable run, worry falls as the told level rises (odds ratio $0.80$ per level, 95\% interval $0.66$ to $0.98$).
\section{Discussion and Limitations}
\label{sec:discussion}
This evaluation shows a sharp dissociation between trait expression and cognitive-diversity fidelity. Across seven simulation runs of four models, agents recovered questionnaire-style NFC and Big Five structure, but fell short on the harder targets: reflection-test scores that track the assigned level, the structure of intuitive-lure errors, the correlation between the two reflection tests, and the variation in decisions written for the BRT scenarios. Claude Sonnet 5 comes closest to people's item-by-item pattern of lures, yet in no run do reflection scores spread across agents as they do across people: scores pile up at the ends of the scale, and even in that run only 20\% of agents score 2 or 3 of 4 on CRT2, against 73\% of people. That spread is the standard a simulation must meet to say something about a real population. This is a problem: ordinary persona prompting can produce locally coherent trait performances on self-report questionnaires while altering the structured variation in behaviour associated with the assigned traits, which is what makes a simulation scientifically informative.

The reflection tests fail in two ways that a single score would hide. In the four Claude Sonnet 4.6 runs, every agent answers every item correctly whatever reflection level it was told, so the assigned level leaves no trace in the test; full marks here are a pull toward correct answers, not a highly reflective persona. In the other three runs lures appear, and the configuration runs show that whether a told reflection level shows as intuitive-lure errors depends first on the model and then, in some models, on whether thinking is turned on (Section~\ref{sec:results_intuitive_lures}). Differences in post-training, such as reinforcement learning for reasoning, are one possible cause; however, these runs do not test it. This fits \citet{hagendorff2023Humanlikeintuitivebehavior}, who found intuitive errors disappearing in later GPT models; here, telling GPT-5.6 Sol its persona's reflection level did not bring them back with reasoning on or off (Section~\ref{sec:results_intuitive_lures}). Where lures do appear, scores on the two CRTs become almost interchangeable, despite correlating only modestly in people (Section~\ref{sec:results_trait_structure}). People find the newly assembled bCRT harder than CRT2 (Appendix Table~\ref{tab:human-crt2-reference}); the agents score much the same on both, so every run sits further from people on the bCRT (Appendix Figure~\ref{fig:distribution_grid}). Together these suggest the told level is applied as a general CRT answering pattern carried over to new items, rather than as a \textit{reflective construct} that interacts with each item as in people.

The strongest implication comes from the contrast between self-report calibration and behavioural compression. If calibrated NFC and BFI responses were sufficient for simulation validity, the BRT should have retained human-like variation in evidence-weighting, epistemic-search and social-obligation tendencies. Instead, every run over-produced the same few behaviours: postponing the medication decision until the claim is checked, consulting authoritative sources, and appealing to fairness or loyalty toward a friend or the restaurant (Section~\ref{sec:results_brt_compression}). Even where a run reproduces one of people's reliable links between bCRT score and a behaviour, the behaviour is over-produced: in Claude Sonnet 5 and Gemini 3.5 Flash, consulting regulators and checking hygiene or official records keep people's direction but not their rate (Figure~\ref{fig:brt_behavioural_compression}). One reading is that this is a stereotype of reflective behaviour rather than its graded human form; another is that it is an assistant default that recommends authoritative sources whatever the persona. However, the fact that some behaviours, such as consulting regulators, rise with the assigned reflection level weighs against a persistent default (Appendix Figure~\ref{fig:brt_level_prevalence}; Appendix~\ref{app:brt_readings}).

These results call for simulation research to specify, and check the realisation of, distributions, error types, associations and downstream behaviour at a granular level. Some models resist the told reflection level almost entirely, but prompt steering does create variation, which leaves room for exploration of prompt designs that improve fidelity. The runs over-produce behaviours but almost never under-produce them (Section~\ref{sec:results_brt_compression}), so the task is less to induce the relevant behaviours than to temper them to the human rate. Finally, we tested only prompt steering; our evaluation suite gives activation steering, reinforcement learning, fine-tuning and other methods a target for higher-fidelity simulation of the decision-making and reasoning associated with psychological constructs.

Several limitations bound our claim. The full battery covers four models; the configuration runs extend only the reasoning items to further models and thinking regimes. The human reference sample comes from the UK and the US, and a more international sample would show whether its targets hold elsewhere. The BRT scenarios are stylised probes rather than exhaustive samples of real-world reasoning, and BRT outcomes rely on LLM-assisted, rubric-based annotation. Future work can address each of these; the central contrast stands: current agent-prompting practice can look trait-conditioned while failing to demonstrate the fine-grained cognitive and behavioural variation required for valid synthetic populations.

\begin{ack}
\paragraph{Funding.} S.W. was supported by a PhD scholarship jointly funded by Aurecon Group and the Commonwealth Scientific and Industrial Research Organisation (CSIRO). We also thank the ARC Centre of Excellence for Automated Decision-Making and Society (CE200100005) for its support. Participant reimbursement for the online (Prolific) study was funded by Aurecon Group. The funders had no role in study design, data collection and analysis, the decision to publish, or preparation of the manuscript.

\paragraph{Competing interests.} A.B. is employed by Aurecon Group. Aurecon Group and the CSIRO had no role in the study design, data collection, analysis, interpretation, or the decision to publish, and this work does not evaluate any product or service of either organisation. The authors declare no other competing interests.
\end{ack}

\bibliographystyle{plainnat}
\bibliography{sn-bibliography}

\clearpage
\appendix

\appendix
\section{Additional Methodological Details}
\label{app:methods_details}

This appendix provides the methodological details needed to interpret and reproduce the cognition-suite evaluation. Its sections follow Sections~\ref{sec:problem_statement} and~\ref{sec:methods}: the evaluation workflow, formal fidelity definitions, human reference data, synthetic profile generation, simulation conditions and configuration runs, response scoring and coding, the LIWC analysis, and reproducibility.

\subsection{Cognition-suite workflow}
\label{app:cognition_suite_workflow}

\begin{figure}[ht]
  \centering
  \includegraphics[width=\linewidth]{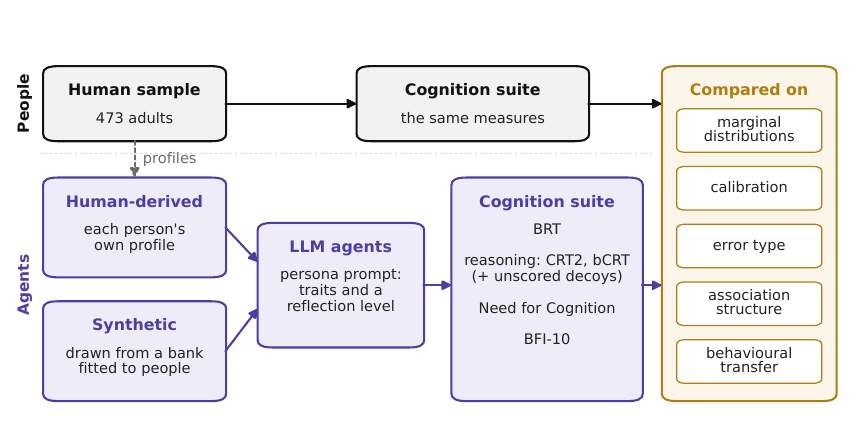}
  \caption{How is each run evaluated? People and LLM agents answer the same cognition suite; the agents are conditioned on profiles taken from people, either each person's own or drawn from a bank fitted to them, and the two sets of answers are compared on the five fidelity targets.}
  \label{fig:cognition_suite_schematic}
\end{figure}

Figure~\ref{fig:cognition_suite_schematic} summarises the evaluation of Sections~\ref{sec:problem_statement} and~\ref{sec:methods}.

\subsection{Formal fidelity definitions}
\label{app:formal_fidelity_definitions}

This section gives the formal definitions behind the fidelity targets of Section~\ref{sec:problem_statement} and the verdicts of Section~\ref{subsec:statistical_analyses}.

Let
\[
H=\{(x_i,y_i,\ell_i)\}_{i=1}^{N}
\]
denote the human reference sample, where $x_i$ contains demographic and psychometric variables, $y_i$ contains coded behavioural outcomes, and $\ell_i$ contains linguistic features from open-ended responses. Let
\[
A_r=\{(\tilde{x}_{jr},\hat{x}_{jr},\hat{y}_{jr},\hat{\ell}_{jr})\}_{j=1}^{M_r}
\]
denote an agent population from run $r$, where $\tilde{x}_{jr}$ is the specified persona profile and $\hat{x}_{jr}$, $\hat{y}_{jr}$, and $\hat{\ell}_{jr}$ are realised psychometric, behavioural, and linguistic outputs.

For marginal fidelity, we compare the human distribution $P_H(v)$ with the realised agent distribution $P_{A_r}(v)$ for each construct $v$ by the Jensen--Shannon divergence (JS):
\[
D_{\mathrm{marg}}(v,r)=\mathrm{JS}\{P_H(v),P_{A_r}(v)\}.
\]
For calibration fidelity, we estimate whether specified agent traits are realised:
\[
\hat{x}_{jrk}=\alpha_{rk}+\beta_{rk}\tilde{x}_{jrk}+\epsilon_{jrk}.
\]
For association fidelity, we compare human and agent Spearman correlation matrices over the shared construct set $V$:
\[
\Delta_r = R_{A_r} - R_H.
\]
For transfer fidelity, we compare human and agent slopes from:
\[
g\{E(o)\}=\alpha+\beta z(x_k)+\gamma^\top c,
\]
where $g$ is the identity link for continuous linguistic outcomes and the logit link for binary behavioural outcomes. Human models use observed traits; agent models use realised traits. For error-type fidelity, Cognitive Reflection Test (CRT) responses are compared over $\{\mathrm{correct}, \mathrm{intuitive}, \mathrm{other}, \mathrm{missing}\}$ response categories.

\paragraph{Behavioural fidelity.} For each of the 15 coded Behavioural Reflection Test (BRT) behaviours $o$, run and human prevalences $\hat{p}_{o,r}$ and $\hat{p}_{o,H}$ carry Wilson 95\% intervals, and the gap $G_{o,r} = \hat{p}_{o,r} - \hat{p}_{o,H}$ (in percentage points) carries a 2,000-resample bootstrap interval. An outcome is \emph{compressed} when $|G_{o,r}| \geq 20$ and the two prevalence intervals do not overlap. Steering is the slope of each outcome on the reflection level the persona was told, per run. Transfer refits each outcome's logistic model on the run's realised bCRT and compares the run slope with the human slope, $\beta_r - \beta_H$, using an independent-samples standard error: the association \emph{transfers} when the interval on the difference includes 0 and the run's own interval excludes 0 in the human direction, \emph{fails to transfer} when the difference interval excludes 0, and is \emph{indeterminate} otherwise. A failure where the human interval includes 0 and the run's own interval excludes 0 means the run \emph{introduces} an association people do not show. A model with fewer than 20 events or fewer than 20 non-events is \emph{unstable}: it is fitted and reported, and left out of the transfer tally. The LIWC slopes (Appendix~\ref{app:liwc_relationship_fidelity}) take the same verdicts with the sub-types of trait-structure transfer below: a failure is weakened, reversed or exaggerated, and a comparison where people show no link and the run introduces none has \emph{no human target}.

\paragraph{Trait-structure transfer.} For each run and each of the 28 pairs of the eight constructs (CRT2, bCRT, Need for Cognition (NFC) and the five traits of the ten-item Big Five Inventory, BFI-10), Spearman correlations (average ranks for ties) are computed in the run, $\rho_r$, and in the human reference, $\rho_H$, with $\Delta\rho = \rho_r - \rho_H$. Each carries a percentile bootstrap 95\% interval from 10,000 resamples (seed 42), with the person as the resampling unit and humans and agents resampled independently; the human frame is resampled once and the same resamples serve every run, the $i$-th human resample paired with the $i$-th agent resample for $\Delta\rho$. A resample in which a score is constant yields no value and is dropped and counted. The count is larger than the 2,000 used for the prevalence gaps because these intervals decide verdict words at their ends. Verdicts are evaluated in this order, the first clause that holds deciding:
\begin{enumerate}[noitemsep]
    \item \emph{not estimable}: a run score is constant;
    \item if the human interval includes 0: \emph{introduced} when the run's interval and the $\Delta\rho$ interval both exclude 0, otherwise \emph{no human target};
    \item \emph{reversed}: the run's interval excludes 0 with the sign opposite to $\rho_H$;
    \item if the $\Delta\rho$ interval excludes 0: \emph{exaggerated} when $|\rho_r| > |\rho_H|$ with the same sign, otherwise \emph{weakened};
    \item \emph{preserved}: the run's interval excludes 0 in the human direction;
    \item \emph{indeterminate} otherwise.
\end{enumerate}
With 471 per side the $\Delta\rho$ interval spans about $\pm.13$ where $\rho$ is near 0 and about $\pm.11$ where $\rho$ is near .4, so \emph{preserved} means within about $\pm.12$ of the human value and non-zero in the human direction: not detectably different at this sample size, not shown equivalent. No correction is applied across pairs, as for transfer, so about one targeted pair in twenty is expected to read as a departure under perfect fidelity.

\subsection{Human reference data}
\label{app:human_reference_data}

This section details the human reference data of Section~\ref{subsec:human_reference_data}. The data were collected online using Qualtrics from adults recruited through Prolific (aged 18 or over, fluent in English, resident in the UK or USA), who were paid at Prolific's recommended ethical hourly rate. The study was approved by the University of New South Wales Human Research Ethics Committee (approval reference iRECS7125); every participant read a Participant Information Statement and gave informed consent in Qualtrics before taking part, was told of the right to withdraw, and was debriefed afterwards. Responses were anonymised at collection and stored securely on the institution's survey platform. Participants completed two BRT writing scenarios, followed by a randomised reasoning block containing CRT2, bCRT, and decoy questions, then Need for Cognition, BFI-10, demographics, and item-familiarity checks. This paper uses the human data as a reference distribution for simulation evaluation; the bCRT and the BRT, and their validation, are presented in the companion study~\citep{weatherhead2026behavioural}. The submitted version of this paper used 467 participants; the companion study's final cleaning reinstated four, giving the 471 used here.

Each participant reported, item by item, whether they had seen it before (Table~\ref{tab:item-exposure}). CRT2 scores were adjusted for this familiarity by a regression of CRT2 on the number of CRT2 items participants reported having seen before, controlling for age; each participant's score was then set to its predicted value at zero items seen (score $-$ $\hat{\beta}$ $\times$ items seen; $\hat{\beta} = 0.167$ points per item). The response-category reference is predicted the same way: for each CRT2 item, a multinomial logit of the response category on the number of CRT2 items seen, age, sex and education, predicted for every participant at zero items seen and averaged (Table~\ref{tab:human-crt2-reference}). The bCRT is not adjusted, because its items were seen far less often (Section~\ref{subsec:human_reference_data}).

\begin{table}[t]
\centering
\footnotesize
\setlength{\tabcolsep}{4pt}
\caption{How familiar were the reflection items, and how much does familiarity lower the human lure rate? Prior exposure per item and the human intuitive-lure rate on all, unseen and seen items. CRT2 items are public and widely seen; the bCRT items were written for the study.}
\label{tab:item-exposure}
\begin{tabularx}{\linewidth}{@{}>{\hsize=1.00\hsize\raggedright\arraybackslash}Xrrp{0.17\linewidth}p{0.17\linewidth}p{0.17\linewidth}r@{}}
\toprule
Item & \begin{tabular}[b]{@{}c@{}}Seen \\ before (\%)\end{tabular} & \begin{tabular}[b]{@{}c@{}}Seen \\ (people)\end{tabular} & \begin{tabular}[b]{@{}c@{}}Lure, all \\ (\%)\end{tabular} & \begin{tabular}[b]{@{}c@{}}Lure, unseen \\ only (\%)\end{tabular} & \begin{tabular}[b]{@{}c@{}}Lure, seen \\ only (\%)\end{tabular} & \begin{tabular}[b]{@{}c@{}}Unseen $-$ \\ all (pp)\end{tabular} \\
\midrule
\multicolumn{7}{@{}p{\linewidth}@{}}{\textit{CRT2}} \\
CRT2 q1 & 74.5 & 351 & 18.7 & 33.3 & 13.7 & +14.6 \\
CRT2 q2 & 58.6 & 276 & 9.8 & 15.9 & 5.4 & +6.1 \\
CRT2 q3 & 74.3 & 350 & 15.3 & 21.5 & 13.1 & +6.2 \\
CRT2 q4 & 40.8 & 192 & 47.6 & 53.4 & 39.1 & +5.8 \\
\textbf{CRT2, all items} & 40.8--74.5 & -- & 22.8 [20.8, 25.1] & 34.4 [30.6, 38.3] & 15.7 [13.4, 18.2] & +11.6 \\
\midrule
\multicolumn{7}{@{}p{\linewidth}@{}}{\textit{bCRT}} \\
bCRT q1 & 31.6 & 149 & 14.2 & 16.8 & 8.7 & +2.5 \\
bCRT q2 & 10.6 & 50 & 56.5 & 59.4 & 32.0 & +2.9 \\
bCRT q3 & 4.9 & 23 & 48.0 & 49.8 & 13.0 & +1.8 \\
bCRT q4 & 5.3 & 25 & 45.6 & 46.4 & 32.0 & +0.8 \\
\textbf{bCRT, all items} & 4.9--31.6 & -- & 41.1 [38.6, 43.6] & 44.8 [42.1, 47.6] & 16.2 [11.2, 21.6] & +3.8 \\
\bottomrule
\end{tabularx}
\par\smallskip
\begin{minipage}{\linewidth}\footnotesize\raggedright
Frame: the 471 CRT2-complete respondents (the human reference of every lure table); N = 471 per item. Seen before = self-report on the item's Experience Check; on the all-items rows, the range across the four items. Lure = intuitive responses / scored responses (none unscored). Brackets: 95\% person-clustered bootstrap intervals (5{,}000 resamples). Unseen and seen are self-selected subgroups, so the unseen-only rate is an upper bound on the exposure effect, not an estimate of it.\par
\end{minipage}
\end{table}

\begin{table}[t]
\centering
\footnotesize
\setlength{\tabcolsep}{4pt}
\caption{What would the human response mix be if no one had seen the items before? The human reference, raw and familiarity-adjusted, per item (\% of scored responses).}
\label{tab:human-crt2-reference}
\begin{tabularx}{\linewidth}{@{}>{\hsize=1.00\hsize\raggedright\arraybackslash}Xrrrrrr@{}}
\toprule
Item & \begin{tabular}[b]{@{}c@{}}Raw \\ correct\end{tabular} & \begin{tabular}[b]{@{}c@{}}Raw \\ lure\end{tabular} & \begin{tabular}[b]{@{}c@{}}Raw \\ other\end{tabular} & \begin{tabular}[b]{@{}c@{}}Adjusted \\ correct\end{tabular} & \begin{tabular}[b]{@{}c@{}}Adjusted \\ lure\end{tabular} & \begin{tabular}[b]{@{}c@{}}Adjusted \\ other\end{tabular} \\
\midrule
\multicolumn{7}{@{}p{\linewidth}@{}}{\textit{CRT2 (familiarity-adjusted reference used for every CRT2 comparison)}} \\
CRT2 q1 & 80.5 & 18.7 & 0.8 & 63.8 & 35.4 & 0.8 \\
CRT2 q2 & 89.8 & 9.8 & 0.4 & 79.3 & 19.8 & 0.9 \\
CRT2 q3 & 84.3 & 15.3 & 0.4 & 76.2 & 22.9 & 0.9 \\
CRT2 q4 & 33.8 & 47.6 & 18.7 & 21.8 & 55.2 & 23.0 \\
\textbf{all four items} & 72.1 & 22.8 & 5.1 & 60.3 & 33.3 & 6.4 \\
\midrule
\multicolumn{7}{@{}p{\linewidth}@{}}{\textit{bCRT (never adjusted: 5--32\% exposure per item)}} \\
bCRT q1 & 60.3 & 14.2 & 25.5 & n/a & n/a & n/a \\
bCRT q2 & 28.7 & 56.5 & 14.9 & n/a & n/a & n/a \\
bCRT q3 & 18.9 & 48.0 & 33.1 & n/a & n/a & n/a \\
bCRT q4 & 36.5 & 45.6 & 17.8 & n/a & n/a & n/a \\
\textbf{all four items} & 36.1 & 41.1 & 22.8 & n/a & n/a & n/a \\
\bottomrule
\end{tabularx}
\par\smallskip
\begin{minipage}{\linewidth}\footnotesize\raggedright
Raw: the 471 CRT2-complete respondents. Adjusted: per-item multinomial logit of the response category on the number of CRT2 items seen before, age, sex and education, predicted at zero items seen and averaged over the 473 respondents of the analytic frame (fitted on the CRT2-complete respondents). n/a: the bCRT is not adjusted.\par
\end{minipage}
\end{table}

\subsection{Synthetic profile generation}
\label{app:persona_generation}

This section details the synthetic profiles of Section~\ref{subsec:persona_sources}. They were generated from the cleaned and scored human reference data. The target schema contained demographics, familiarity-adjusted CRT2 (Appendix~\ref{app:human_reference_data}), bCRT (sampled with the rest but never given to a persona), Need for Cognition, and the five BFI traits. The bCRT is never given because a level on it would refer to nothing a model could know: its puzzles may circulate online, but not as a CRT with published human data, so a model has no record of, for example, how often people give each lure. CRT2, a published test with published human data, carries the told level instead.

Numeric cognitive and personality variables were sampled using a Gaussian copula. Variables were rank-Gaussianised, a latent correlation matrix was estimated, new samples were drawn from the corresponding multivariate normal distribution, and values were mapped back to the observed scale. Continuous variables used interpolated empirical inverse CDFs; count-like variables retained discrete support.

Categorical demographic variables were sampled using targeted conditional distributions. Education and employment were sampled conditional on age band. Occupation was sampled conditional on employment status, education, and age band, with hierarchical fallback to simpler conditionals when cells were sparse. Missing education and occupation labels were excluded from the synthetic target distribution.

The resulting bank contains 5,000 synthetic profiles; its mean familiarity-adjusted CRT2 score is within 0.005 points of people's (Jensen--Shannon divergence 0.0004). This second release is the one the seven simulation runs draw on; the configuration-run panel (Appendix~\ref{app:prompt_screening}) comes from the first release, whose mean was 0.23 points below people's.

The CRT2 level cutoffs (Section~\ref{subsec:persona_sources}) and the profile bank were estimated from the reference data before the final scoring pass, which credited six participants' correct item-4 responses (`none'); both are released as used. The specified CRT2 distribution of the 471 synthetic profiles differs marginally from the final reference (Jensen--Shannon divergence 0.0023, against 0.42--0.73 between agents and people), and ties on the lowest cutoff were assigned to level 1.

\subsection{Simulation conditions}
\label{app:simulation_conditions}

Table~\ref{tab:experimental-runs} lists the seven full-battery simulation runs of Section~\ref{subsec:agent_simulation_protocol}.

\begin{table}[t]
\centering
\footnotesize
\setlength{\tabcolsep}{4pt}
\caption{Experimental conditions: one row per full-battery simulation run.}
\label{tab:experimental-runs}
\begin{tabularx}{\linewidth}{@{}>{\hsize=1.15\hsize\raggedright\arraybackslash}X>{\hsize=0.90\hsize\raggedright\arraybackslash}Xl>{\hsize=1.00\hsize\raggedright\arraybackslash}X>{\hsize=0.95\hsize\raggedright\arraybackslash}Xr@{}}
\toprule
Condition & Persona source & Format & Model & Configuration & N \\
\midrule
Claude Sonnet 5 & Synthetic profiles & Numeric & Claude Sonnet 5 (claude-sonnet-5) & provider defaults; adaptive thinking on & 471 \\
Gemini 3.5 Flash & Synthetic profiles & Numeric & Gemini 3.5 Flash (gemini-3.5-flash) & thinking level minimal & 471 \\
DeepSeek V4-Pro & Synthetic profiles & Numeric & DeepSeek V4-Pro (via Together AI) & reasoning effort high & 471 / 467\textsuperscript{*} \\
Claude Sonnet 4.6 · human · numeric & Human-derived profiles & Numeric & Claude Sonnet 4.6 (claude-sonnet-4-6) & provider defaults & 471 \\
Claude Sonnet 4.6 · human · lexical & Human-derived profiles & Lexical & Claude Sonnet 4.6 (claude-sonnet-4-6) & provider defaults & 471 \\
Claude Sonnet 4.6 · synthetic · numeric & Synthetic profiles & Numeric & Claude Sonnet 4.6 (claude-sonnet-4-6) & provider defaults & 471 \\
Claude Sonnet 4.6 · synthetic · lexical & Synthetic profiles & Lexical & Claude Sonnet 4.6 (claude-sonnet-4-6) & provider defaults & 471 \\
\bottomrule
\end{tabularx}
\par\smallskip
\begin{minipage}{\linewidth}\footnotesize\raggedright
One model call per block (BRT, reasoning, Need for Cognition, BFI), with conversation history retained within each persona's session; CRT2, bCRT and decoy items in randomised order; numeric = explicit scale values, lexical = graded verbal descriptors; seed 42. N = simulated personas (CRT / BRT where they differ). \textsuperscript{*}DeepSeek V4-Pro BRT N = 467: four personas returned no usable open-ended response, and the provider has since retired the model snapshot. Run identifiers: Table~\ref{tab:run-identifiers}.\par
\end{minipage}
\end{table}

\subsection{Configuration runs and prompt format}
\label{app:prompt_screening}

This section details the configuration runs and the choice of prompt format in Section~\ref{subsec:agent_simulation_protocol}. Configuration runs give the reasoning block first, then Need for Cognition and the BFI-10, with no BRT, to one fixed panel of 150 synthetic profiles, 30 at each reflection level, drawn from the first release of the profile bank (seed 42); the paper uses their eight CRT2 and bCRT items. One shared panel means every configuration answers as the same profiles, so differences between configurations are not sample differences, and each low-reflection level, where lures are diagnostic, has 30 profiles, which a random draw does not guarantee. The full 471-profile battery is reserved for the configuration in which each model is evaluated. Table~\ref{tab:configuration-ladder} gives the lure rate of every configuration.

Persona prompting commonly uses both textual and numerical attributes, but, to the best of our knowledge, prior work has not definitively established how these formats compare for preserving psychologically meaningful simulation fidelity across calibration, association structure, and downstream behaviour \cite{salminen2025UsingAIUser}. We therefore treat numeric and lexical prompting as two plausible specification formats rather than assuming either is superior. Lexical specification also follows psychometrically grounded personality-shaping work using verbal trait descriptors \cite{serapio-garcia2025psychometricframeworkevaluating}.

Prompt format was chosen in two steps. Before the full runs, a 100-persona screen of Claude Sonnet 4.6 compared numeric and lexical profiles on the reasoning, Need for Cognition and BFI-10 blocks, and numeric gave stronger questionnaire fidelity; Claude Sonnet 4.6 was then run in both formats, so no format choice gates its results. Because Claude Sonnet 4.6 produces no lures, the format was checked again for Gemini 3.5 Flash and DeepSeek V4-Pro by a criterion set before the runs: a lexical configuration run of each was set against its numeric run, the format whose pooled CRT2 and bCRT response categories lay closer to the familiarity-adjusted human reference (lower Jensen--Shannon divergence) would be used, and a difference under 0.02 would go to numeric. Numeric was chosen for both: 0.231 against 0.279 for Gemini 3.5 Flash, and 0.324 against 0.334 for DeepSeek V4-Pro. Claude Sonnet 5 kept the numeric default.

Table~\ref{tab:design-probes} gives three checks on the absence of lures in Claude Sonnet 4.6. Decoding temperature and the retained conversation history (Section~\ref{subsec:agent_simulation_protocol}) change nothing: no setting produces a lure. Only a framing that tells the persona plausible wrong answers are correct role performance produces lures, in 10.2\% of answers.

\begin{table}[t]
\centering
\footnotesize
\setlength{\tabcolsep}{4pt}
\caption{Which configurations produce intuitive-lure errors? The configuration ladder on the reasoning core: every model and thinking regime, each answering the eight reflection items for the same 150 profiles.}
\label{tab:configuration-ladder}
\begin{tabularx}{\linewidth}{@{}>{\hsize=1.00\hsize\raggedright\arraybackslash}X>{\hsize=1.00\hsize\raggedright\arraybackslash}Xp{0.17\linewidth}p{0.17\linewidth}p{0.17\linewidth}@{}}
\toprule
Model & Configuration & \begin{tabular}[b]{@{}c@{}}CRT2 lure \\ \% [95\% CI]\end{tabular} & \begin{tabular}[b]{@{}c@{}}bCRT lure \\ \% [95\% CI]\end{tabular} & \begin{tabular}[b]{@{}c@{}}Pooled lure \\ \% [95\% CI]\end{tabular} \\
\midrule
\multicolumn{5}{@{}p{\linewidth}@{}}{\textit{Anthropic}} \\
Claude Sonnet 5 & adaptive thinking (default) & 20.8 [15.8, 26.3] & 13.8 [10.0, 17.8] & \textbf{17.3} [13.2, 21.8] \\
Claude Sonnet 5 & thinking disabled & 4.5 [3.0, 6.0] & 0.2 [0.0, 0.5] & \textbf{2.3} [1.6, 3.2] \\
Claude Fable 5 & provider defaults & 32.0 [25.3, 39.0] & 40.3 [34.7, 45.8] & \textbf{36.2} [30.4, 42.1] \\
\midrule
\multicolumn{5}{@{}p{\linewidth}@{}}{\textit{Google}} \\
Gemini 3.5 Flash & thinking level minimal & 23.0 [16.8, 29.7] & 19.7 [14.8, 25.0] & \textbf{21.3} [15.9, 27.2] \\
Gemini 3.5 Flash & thinking level minimal (lexical) & 17.7 [12.2, 23.7] & 14.7 [10.2, 19.3] & \textbf{16.2} [11.2, 21.4] \\
\midrule
\multicolumn{5}{@{}p{\linewidth}@{}}{\textit{DeepSeek}} \\
DeepSeek V4-Pro, first-party API & thinking enabled & 9.6 [5.5, 14.1] & 12.1 [7.2, 17.4] & \textbf{10.8} [6.5, 15.6] \\
DeepSeek V4-Pro, first-party API & thinking enabled (lexical) & 9.8 [6.0, 14.2] & 12.2 [7.7, 17.0] & \textbf{11.0} [6.9, 15.4] \\
DeepSeek V4-Pro, Together AI & reasoning effort high & 11.5 [7.2, 16.3] & 13.8 [8.8, 19.3] & \textbf{12.7} [8.0, 17.8] \\
DeepSeek V4-Pro, Together AI & reasoning disabled & 0.3 [0.0, 0.8] & 1.8 [0.8, 3.0] & \textbf{1.1} [0.6, 1.7] \\
DeepSeek V4-Flash & thinking enabled & 1.5 [0.0, 3.5] & 1.3 [0.2, 2.7] & \textbf{1.4} [0.2, 3.0] \\
DeepSeek V4-Flash & thinking disabled & 0.0 (0/600; no lures observed) & 2.3 [1.2, 3.7] & \textbf{1.2} [0.6, 1.8] \\
\midrule
\multicolumn{5}{@{}p{\linewidth}@{}}{\textit{OpenAI}} \\
GPT-5.6 Sol & reasoning effort none & 0.0 (0/600; no lures observed) & 0.0 (0/600; no lures observed) & \textbf{0.0} (0/1200; no lures observed) \\
GPT-5.6 Sol & reasoning effort high & 0.0 (0/600; no lures observed) & 0.2 [0.0, 0.5] & \textbf{0.1} [0.0, 0.2] \\
\midrule
\multicolumn{5}{@{}p{\linewidth}@{}}{\textit{Moonshot AI}} \\
Kimi K3, Together AI & provider-default effort & 6.0 [2.7, 9.8] & 6.2 [2.8, 10.0] & \textbf{6.1} [2.8, 9.8] \\
\bottomrule
\end{tabularx}
\par\smallskip
\begin{minipage}{\linewidth}\footnotesize\raggedright
N = 150 profiles per configuration (1{,}200 item responses; 600 per test): a fixed, stratified panel of 150 synthetic profiles (30 per specified reflection level) drawn from the first release of the profile bank and its CRT2 level cutoffs, released with the paper (\texttt{CONFIG\_PANEL\_150.csv}). Lure = intuitive responses / scored responses. Intervals are 95\% persona-clustered bootstrap intervals (5{,}000 resamples); a cell with no lures gives its count and ``no lures observed'' in place of an interval. Unscored (blank) answers, excluded from the denominator of both the rate and its interval: DeepSeek V4-Pro, first-party API (thinking enabled): 4 CRT2 and 4 bCRT. Numeric profile format unless marked lexical.\par
\end{minipage}
\end{table}

\begin{table}[t]
\centering
\footnotesize
\setlength{\tabcolsep}{4pt}
\caption{Do decoding temperature, the retained conversation history or a sanction framing explain the absence of lures in Claude Sonnet 4.6? Three design checks on the same profile panel.}
\label{tab:design-probes}
\begin{tabularx}{\linewidth}{@{}>{\hsize=0.80\hsize\raggedright\arraybackslash}Xrrrp{0.17\linewidth}>{\hsize=1.20\hsize\raggedright\arraybackslash}X@{}}
\toprule
Cell & N & \begin{tabular}[b]{@{}c@{}}CRT2 \\ lure \%\end{tabular} & \begin{tabular}[b]{@{}c@{}}bCRT \\ lure \%\end{tabular} & \begin{tabular}[b]{@{}c@{}}Pooled lure \\ \% [95\% CI]\end{tabular} & Check \\
\midrule
\multicolumn{6}{@{}p{\linewidth}@{}}{\textit{Temperature sweep}} \\
temperature 0.0 & 150 & 0.0 & 0.0 & 0.0 [0.0, 0.0] & NFC variance 103\% of human \\
temperature 0.3 & 150 & 0.0 & 0.0 & 0.0 [0.0, 0.0] & NFC variance 103\% of human \\
temperature 0.7 & 150 & 0.0 & 0.0 & 0.0 [0.0, 0.0] & NFC variance 107\% of human \\
temperature 1.0 & 150 & 0.0 & 0.0 & 0.0 [0.0, 0.0] & NFC variance 105\% of human \\
temperature provider default (1.0) & 150 & 0.0 & 0.0 & 0.0 [0.0, 0.0] & NFC variance 104\% of human \\
\midrule
\multicolumn{6}{@{}p{\linewidth}@{}}{\textit{No-history ablation}} \\
with history (original run) & 150 & 0.0 & 0.0 & 0.0  &  \\
no history & 150 & 0.0 & 0.0 & 0.0  & paired with the arm above: McNemar exact p = 1.00; 150/150 concordant \\
\midrule
\multicolumn{6}{@{}p{\linewidth}@{}}{\textit{Sanction probe}} \\
standard framing & 100 & 0.0 & 0.0 & 0.0 [0.0, 0.0] &  \\
sanction framing & 100 & 12.8 & 7.8 & 10.2 [7.8, 13.1] & 50/100 personas produce a lure only under sanction; McNemar exact p = $1.8\times 10^{-15}$ \\
\bottomrule
\end{tabularx}
\par\smallskip
\begin{minipage}{\linewidth}\footnotesize\raggedright
Claude Sonnet 4.6, numeric profiles. Temperature sweep and no-history ablation: the 150-profile panel of the configuration ladder; the with-history arm is the same 150 profiles in a full-battery numeric run of Claude Sonnet 4.6 on the first release of the profile bank, the no-history arm re-administers the reasoning block with no prior conversation. Sanction probe: 100 profiles (50 at each of told CRT2 levels 0 and 1); the sanction framing appends ``plausible-but-wrong intuitive answers are correct role performance, not errors''. Lure = intuitive responses / scored responses; intervals are 95\% persona-clustered bootstrap intervals. NFC variance is the realised NFC variance as a percentage of the human NFC variance.\par
\end{minipage}
\end{table}

\subsection{Response scoring}
\label{app:crt_response_categories}

This section details the scoring of Section~\ref{subsec:human_reference_data}. CRT2, bCRT, and decoy responses were scored in two stages, for people and agents alike: each answer was normalised and matched against item-specific answer keys, which list the correct answer and the intuitive lure, and answers the keys did not resolve were classified by three large language models (LLMs) from different providers (GPT-5.6 Sol, Gemini 3.5 Flash and DeepSeek V4-Flash), and the majority label was kept. Every vote is released. Against a blinded annotation of 125 human answers by one author, the first version of this scoring agreed on 86.4\% of answers ($\kappa = .79$); the released version agrees with the same annotation, after the annotation's own errors were corrected, on 99.2\%, a consistency check rather than an independent validation.

CRT2, bCRT, and decoy responses were assigned one of four categories:
\[
\{\mathrm{correct}, \mathrm{intuitive}, \mathrm{other}, \mathrm{missing}\},
\]
where missing is a blank answer. Lure rates are intuitive-lure responses divided by scored responses, which leave out missing answers; for lure rates, the human reference is the 471 CRT2-complete participants. Person-level totals were computed separately for correct, intuitive-lure, and other responses. Need for Cognition and BFI-10 were scored deterministically from item responses, using standard reverse-scoring rules.

\subsection{BRT coding rubric}
\label{app:brt_coding_rubric}

This section gives the BRT rubric behind the 15 coded behaviours of Section~\ref{subsec:human_reference_data}. BRT responses were coded with one rubric, the same for people and agents and released with the code, by three LLMs from different providers (GPT-5.6 Sol with low reasoning effort, Gemini 3.5 Flash with minimal thinking, and DeepSeek V4-Flash), with the majority label kept. Two coding passes over the same 150 responses measured the panel's agreement field by field, against a gate of 95\% for the two decisions and what each rested on, and 90\% for the other fields: 25 of the 28 coded-field gates pass. The three below their gate are reported with their agreement: what the restaurant decision rested on (85.3\%, $\kappa = .79$), the medication decision (90.0\%, $\kappa = .85$), and what the medication decision rested on (79.3\%, $\kappa = .73$).

Each of the 15 coded behaviours is present or absent in a response. \emph{Sources the respondent plans to consult} (seven): phoning the restaurant; regulators, journals or official guidance; friends or family about the medication; hygiene or official records; a qualified professional; friends or family about the restaurant; and the sources the video cites. \emph{Decision taken} (five, from the coded decision and its stated reasons): keeps the booking on evidence grounds (keeps it, and gives a hygiene or risk reason, plans to check official records, or rests the decision on the critics' or official record); accepts on institutional grounds with a verification plan (takes the medication, rests the decision on the medical system, plans a check that names evidence or a professional, and voices no conspiracy idea); delays with a verification plan and a stated condition (postpones the decision, plans such a check, and states when they would take it); acts on the claim with no doubt and no rule (delays or refuses without doubting the video and without a stated condition); and refuses outright, worry-marked (refuses, voices worry, and states no condition). \emph{Emotion expressed} (two): worry about the medication, and hostile rebuke of the unhelpful friend (Friend A in the scenario). \emph{Appeals to fairness or loyalty} (one): toward the friend whose birthday dinner it is (Friend B in the scenario; ``the birthday friend'' in the tables), or toward the restaurant. Ambiguous responses were assigned conservative labels.

\subsection{LIWC analysis}
\label{app:liwc_analysis}

This section details the supplementary LIWC analysis of Section~\ref{subsec:human_reference_data}; its results are in Appendix~\ref{app:liwc_relationship_fidelity}. Linguistic Inquiry and Word Count (LIWC) was applied to the concatenated BRT responses, producing one row per participant or agent. LIWC categories were grouped into theory-relevant families, including quantity/structure, cognitive processes, auxiliary verbs, risk/time-focus, affect, and normative content. The focal LIWC set used in the supplementary relationship-fidelity analysis included word count, negative emotion, differentiation, auxiliary verbs, risk, present-focus, and moral language. These categories were selected because they are theoretically relevant to the BRT design and prior reflection-language work \cite{mosleh2021}. Full category-level regressions use cognitive-reflection predictors and apply Benjamini--Hochberg correction within LIWC families. For relationship fidelity, each run's agents and people were pooled in one ordinary least-squares model per focal category and test: LIWC score on test score, source, and their interaction. A slope is the change in LIWC score per test point; the interaction tests the difference between the run's slope and people's, at $p < .05$ without correction across categories. People's predictors are the familiarity-adjusted CRT2 score (Appendix~\ref{app:human_reference_data}) and the bCRT score as answered; agents' are their own scores on the two CRTs. Where every agent in a run has the same score, no slope can be estimated.

\subsection{Reproducibility}
\label{app:reproducibility}

\begin{table}[t]
\centering
\footnotesize
\setlength{\tabcolsep}{4pt}
\caption{Run identifiers in the released code and data.}
\label{tab:run-identifiers}
\begin{tabularx}{\linewidth}{@{}>{\hsize=1.00\hsize\raggedright\arraybackslash}Xl@{}}
\toprule
Condition & Run identifier \\
\midrule
Claude Sonnet 5 & \texttt{CR\_\allowbreak MAIN\_\allowbreak SYNTH\_\allowbreak NUM\_\allowbreak SONNET5\_\allowbreak S42} \\
Gemini 3.5 Flash & \texttt{CR\_\allowbreak MAIN\_\allowbreak SYNTH\_\allowbreak NUM\_\allowbreak GEM35F\_\allowbreak MIN\_\allowbreak S42} \\
DeepSeek V4-Pro & \texttt{CR\_\allowbreak MAIN\_\allowbreak SYNTH\_\allowbreak NUM\_\allowbreak DSV4PRO\_\allowbreak TOG\_\allowbreak S42} \\
Claude Sonnet 4.6 · human · numeric & \texttt{CR\_\allowbreak BASELINE\_\allowbreak HUMAN\_\allowbreak NUMERIC\_\allowbreak CS46\_\allowbreak S42} \\
Claude Sonnet 4.6 · human · lexical & \texttt{CR\_\allowbreak BASELINE\_\allowbreak HUMAN\_\allowbreak LEXICAL\_\allowbreak CS46\_\allowbreak S42} \\
Claude Sonnet 4.6 · synthetic · numeric & \texttt{CR\_\allowbreak MAIN\_\allowbreak SYNTH\_\allowbreak NUMERIC\_\allowbreak CS46\_\allowbreak S42} \\
Claude Sonnet 4.6 · synthetic · lexical & \texttt{CR\_\allowbreak MAIN\_\allowbreak SYNTH\_\allowbreak LEXICAL\_\allowbreak CS46\_\allowbreak S42} \\
\bottomrule
\end{tabularx}
\par\smallskip
\begin{minipage}{\linewidth}\footnotesize\raggedright
Each analysis report is named \texttt{cr13\_} followed by the run identifier.\par
\end{minipage}
\end{table}

This section gives what is needed to rerun the evaluation. Table~\ref{tab:run-identifiers} gives the identifiers of every run.

\paragraph{Model configuration.} Every run left sampling at the provider's defaults: temperature and top-$p$ were omitted from every call. Claude Sonnet 4.6 (without extended thinking) and Claude Sonnet 5 (adaptive thinking, the provider default) were called through the Anthropic Messages API with output limits of 8,000 and 16,000 tokens; Gemini 3.5 Flash (thinking level minimal) through Google's OpenAI-compatible endpoint with a limit of 8,000; and DeepSeek V4-Pro (reasoning effort high) through Together AI with a limit of 40,000. Every run used seed 42, one call per block with the persona's conversation history retained, and the reasoning items in randomised order. The default configuration was chosen deliberately: deterministic decoding suppresses by construction the response variance the evaluation measures, and the default is what a deployment inherits unless the practitioner intervenes. A temperature sweep on Claude Sonnet 4.6 changes nothing (Table~\ref{tab:design-probes}).

\paragraph{Release.} The release (code: \url{https://github.com/sion-w/CogSuite}, archived at \url{https://doi.org/10.5281/zenodo.24094423}; data: \url{https://osf.io/h8d5s/}) includes aggregate tables for every human reference distribution; the synthetic profile bank as used, with its generation code; the complete answer keys and scoring code; the BRT rubric with worked coding examples; and the configuration files, run metadata and scripts that regenerate every figure and table. Agent-simulation code was generated using the SOCIA-EVO simulator framework \cite{hua2026SOCIAEVOAutomatedSimulator}. Raw human free-text and individual-level human data are not part of this release. Participants consented to their information being shared only in a format that does not identify them, for future research only within the permissions each gave, and to anonymised data being released publicly. De-identified individual-level records are available from the authors now and will be released publicly with the companion study's updated dataset~\citep{weatherhead2026behavioural}.

\FloatBarrier
\section{Additional Results and Diagnostics}
\label{app:additional_results}

The following diagnostics supplement the main-text results and follow Section~\ref{sec:results}: marginal distributions, calibration, lure errors by told level, trait-structure transfer, BRT behaviour, and LIWC relationship fidelity.

\subsection{Additional marginal distribution diagnostics}
\label{app:marginal_distribution_diagnostics}

\begin{figure}[p]
  \centering
  \includegraphics[width=\linewidth]{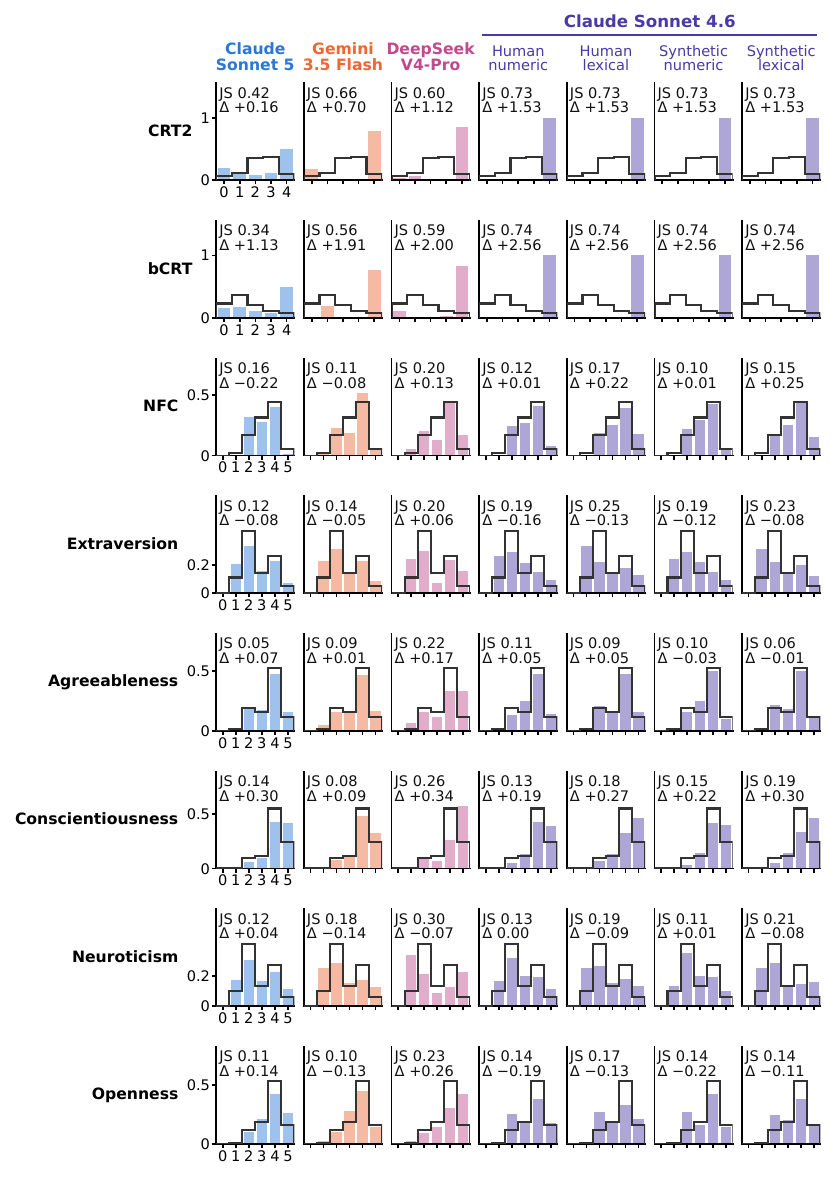}
  \caption{Does each run reproduce the shape of the human score distribution, construct by construct? Rows are the eight constructs and columns the seven full-battery runs, grouped by model family. Bars show each run's score distribution in its family hue; the black outline is the human distribution, repeated across the row, and the $y$-axis is shared within a row. Each panel's corner gives the Jensen--Shannon divergence from the human distribution (JS) and the mean shift $\Delta$ (run $-$ human). $N = 471$ per run; human $N = 471$ (bCRT 473).}
  \label{fig:distribution_grid}
\end{figure}

Figure~\ref{fig:distribution_grid} extends Figure~\ref{fig:distribution_fidelity} to all eight constructs, with each Claude Sonnet 4.6 condition in its own column.

\FloatBarrier
\subsection{Specified-to-realised calibration diagnostics}
\label{app:calibration_diagnostics}

\begin{figure}[ht]
  \centering
  \includegraphics[width=\linewidth]{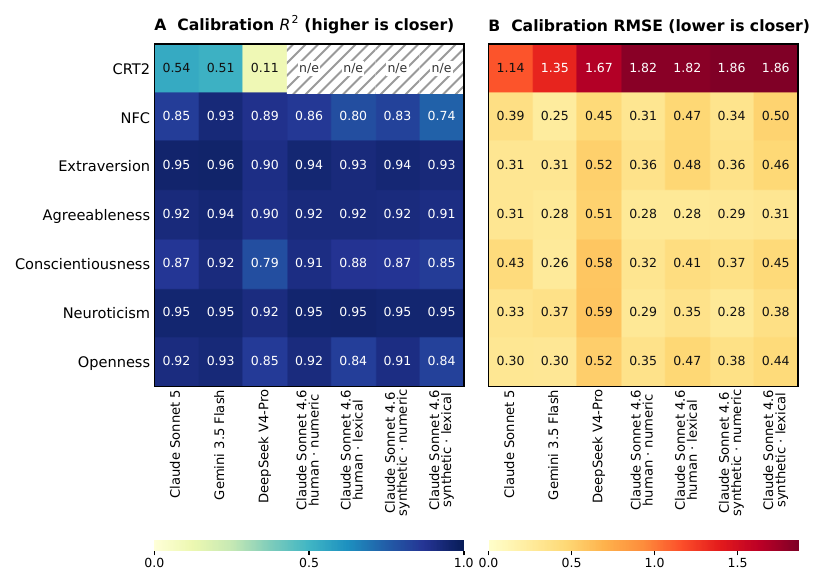}
  \caption{How closely does each run's realised score follow its persona's specified score, construct by construct? (A) $R^2$ and (B) RMSE of the realised score regressed on the specified score, one column per run. n/e: every agent scored 4 of 4, so $R^2$ cannot be estimated.}
  \label{fig:calibration_fingerprint}
\end{figure}

Figure~\ref{fig:calibration_fingerprint} gives calibration for every specified construct and run, behind the finding that self-reported traits follow their specification more closely than cognitive reflection (Section~\ref{sec:results_distribution_fidelity}).

\FloatBarrier
\subsection{Lure errors by told reflection level}
\label{app:crt_error_categories_by_level}

\begin{figure}[ht]
  \centering
  \includegraphics[width=\linewidth]{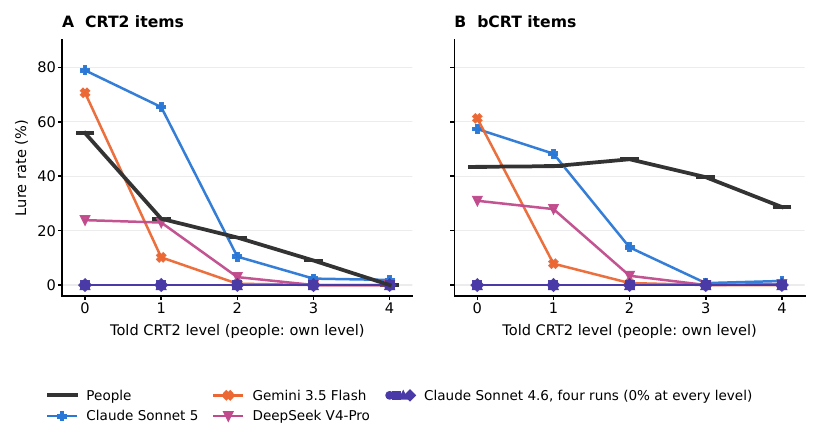}
  \caption{Does telling a persona a lower reflection level produce more lure errors? Lure rate on CRT2 (A) and bCRT (B) at each told CRT2 level, 0--4, in each run, and in people (black) by their own CRT2 level, at the cutoffs the told levels were built on. The bCRT was never specified: its panel groups the same agents by their told CRT2 level. $n$ at levels 0--4: synthetic-profile runs 109/86/103/106/67, human-derived runs 94/98/106/105/68, people 107/79/106/111/68.}
  \label{fig:appendix_crt_response_categories_by_crt_level}
\end{figure}

Figure~\ref{fig:appendix_crt_response_categories_by_crt_level} backs the lure results of Section~\ref{sec:results_intuitive_lures}: it gives the lure rate at each reflection level the persona was told, against people's rate at their own CRT2 level. This checks whether low-reflection specifications induce the human-like error they are meant to induce, rather than merely reducing accuracy. In the three runs that produce lures, lure rates fall steeply with the told level on both tests. On the bCRT, which was never specified, people's rate falls far less across their own CRT2 levels, and from level 2 up the runs sit far below it; the four Claude Sonnet 4.6 runs give none at any level.

\FloatBarrier
\subsection{Trait-structure transfer}
\label{app:trait_structure_transfer}

\begin{figure}[p]
  \centering
  \includegraphics[width=\linewidth]{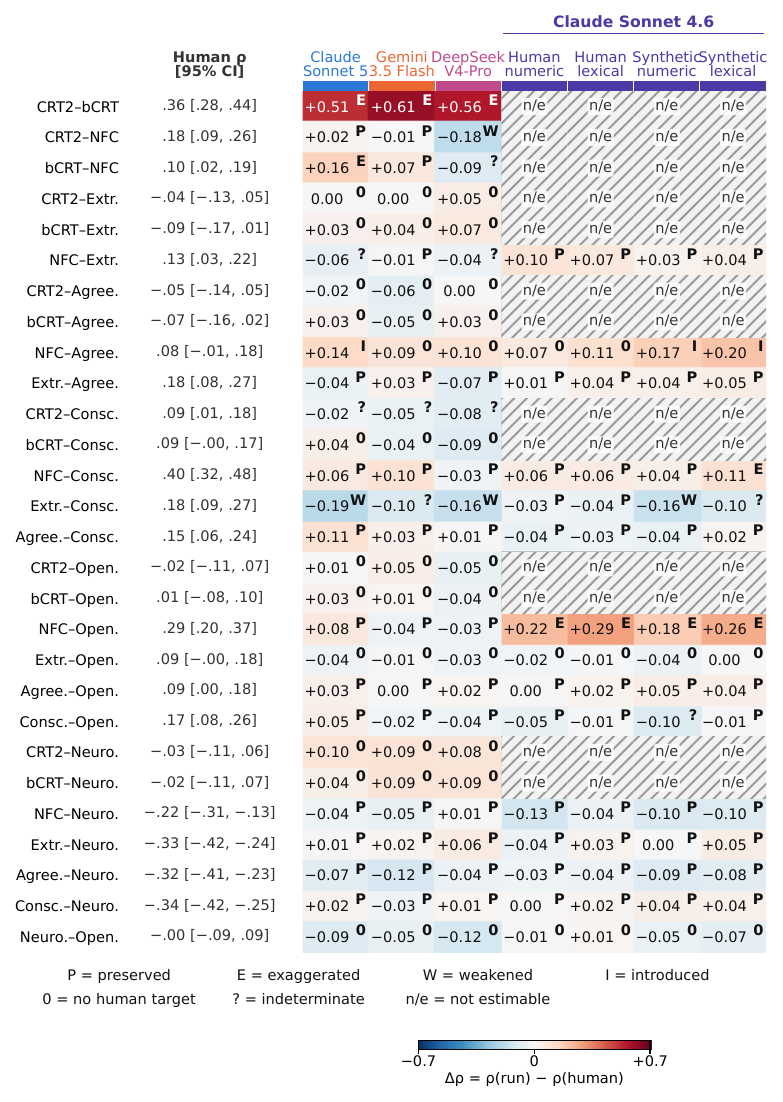}
  \caption{Does each run reproduce the human correlation between every pair of constructs? Rows are the 28 pairs (Extr.\ Extraversion, Agree.\ Agreeableness, Consc.\ Conscientiousness, Neuro.\ Neuroticism, Open.\ Openness); the first column gives the human Spearman correlation with its 95\% interval, each cell the run's difference from it, $\Delta\rho = \rho_{\mathrm{run}} - \rho_{\mathrm{human}}$, with the verdict symbols of Figure~\ref{fig:association_fidelity}; I (introduced) marks a pair with no human correlation but a clearly non-zero run correlation; n/e, every agent scored 4/4. $N = 471$ per side; 95\% percentile bootstrap intervals from 10,000 resamples; no correction across pairs.}
  \label{fig:f4_all_pairs}
\end{figure}

Figure~\ref{fig:f4_all_pairs} and Tables~\ref{tab:f4-a} and~\ref{tab:f4-b} give the trait-structure verdicts for all 28 construct pairs, behind Section~\ref{sec:results_trait_structure}.

\begin{sidewaystable}
\centering
\footnotesize
\setlength{\tabcolsep}{3.5pt}
\caption{Trait-structure transfer, Claude Sonnet 5, Gemini 3.5 Flash and DeepSeek V4-Pro: per construct pair, the run's correlation, its difference from the human correlation, and the verdict.}
\label{tab:f4-a}
\begin{tabular}{@{}llrlcrlcrlc@{}}
\toprule
 &  & \multicolumn{3}{c}{Claude Sonnet 5} & \multicolumn{3}{c}{Gemini 3.5 Flash} & \multicolumn{3}{c}{DeepSeek V4-Pro} \\
\cmidrule(lr){3-5} \cmidrule(lr){6-8} \cmidrule(lr){9-11}
Pair & Human $\rho$ [95\% CI] & $\rho$ & $\Delta\rho$ [95\% CI] & V & $\rho$ & $\Delta\rho$ [95\% CI] & V & $\rho$ & $\Delta\rho$ [95\% CI] & V \\
\midrule
CRT2--bCRT & .36 [.28, .44] & .88 & +0.51 [+0.43, +0.60] & E & .97 & +0.61 [+0.53, +0.70] & E & .92 & +0.56 [+0.47, +0.65] & E \\
CRT2--NFC & .18 [.09, .26] & .20 & +0.02 [$-$0.10, +0.15] & P & .17 & $-$0.01 [$-$0.13, +0.12] & P & .00 & $-$0.18 [$-$0.30, $-$0.05] & W \\
CRT2--Conscientiousness & .09 [.01, .18] & .07 & $-$0.02 [$-$0.15, +0.10] & ? & .04 & $-$0.05 [$-$0.18, +0.08] & ? & .01 & $-$0.08 [$-$0.21, +0.05] & ? \\
bCRT--NFC & .10 [.02, .19] & .27 & +0.16 [+0.04, +0.28] & E & .18 & +0.07 [$-$0.06, +0.20] & P & .01 & $-$0.09 [$-$0.22, +0.03] & ? \\
bCRT--Conscientiousness & .09 [$-$.00, .17] & .13 & +0.04 [$-$0.08, +0.17] & 0 & .04 & $-$0.04 [$-$0.17, +0.09] & 0 & $-$.01 & $-$0.09 [$-$0.22, +0.04] & 0 \\
NFC--Conscientiousness & .40 [.32, .48] & .46 & +0.06 [$-$0.05, +0.17] & P & .50 & +0.10 [$-$0.01, +0.20] & P & .37 & $-$0.03 [$-$0.14, +0.08] & P \\
CRT2--Extraversion & $-$.04 [$-$.13, .05] & $-$.04 & $-$0.00 [$-$0.13, +0.12] & 0 & $-$.04 & $-$0.00 [$-$0.13, +0.13] & 0 & .02 & +0.05 [$-$0.08, +0.18] & 0 \\
bCRT--Extraversion & $-$.09 [$-$.17, .01] & $-$.05 & +0.03 [$-$0.09, +0.16] & 0 & $-$.04 & +0.04 [$-$0.08, +0.17] & 0 & $-$.01 & +0.07 [$-$0.06, +0.20] & 0 \\
NFC--Extraversion & .13 [.03, .22] & .07 & $-$0.06 [$-$0.19, +0.08] & ? & .12 & $-$0.01 [$-$0.14, +0.12] & P & .08 & $-$0.04 [$-$0.17, +0.09] & ? \\
CRT2--Agreeableness & $-$.05 [$-$.14, .05] & $-$.07 & $-$0.02 [$-$0.15, +0.11] & 0 & $-$.11 & $-$0.06 [$-$0.19, +0.07] & 0 & $-$.04 & +0.00 [$-$0.12, +0.13] & 0 \\
bCRT--Agreeableness & $-$.07 [$-$.16, .02] & $-$.04 & +0.03 [$-$0.10, +0.16] & 0 & $-$.13 & $-$0.05 [$-$0.18, +0.07] & 0 & $-$.04 & +0.03 [$-$0.10, +0.15] & 0 \\
NFC--Agreeableness & .08 [$-$.01, .18] & .22 & +0.14 [+0.01, +0.26] & I & .18 & +0.09 [$-$0.04, +0.22] & 0 & .18 & +0.10 [$-$0.03, +0.23] & 0 \\
Extraversion--Agreeableness & .18 [.08, .27] & .13 & $-$0.04 [$-$0.18, +0.09] & P & .20 & +0.03 [$-$0.10, +0.16] & P & .10 & $-$0.07 [$-$0.20, +0.06] & P \\
Extraversion--Conscientiousness & .18 [.09, .27] & $-$.01 & $-$0.19 [$-$0.32, $-$0.06] & W & .08 & $-$0.10 [$-$0.23, +0.02] & ? & .02 & $-$0.16 [$-$0.29, $-$0.03] & W \\
Agreeableness--Conscientiousness & .15 [.06, .24] & .26 & +0.11 [$-$0.02, +0.23] & P & .19 & +0.03 [$-$0.10, +0.16] & P & .16 & +0.01 [$-$0.12, +0.14] & P \\
CRT2--Openness & $-$.02 [$-$.11, .07] & $-$.02 & +0.01 [$-$0.12, +0.13] & 0 & .02 & +0.05 [$-$0.08, +0.17] & 0 & $-$.07 & $-$0.05 [$-$0.17, +0.08] & 0 \\
bCRT--Openness & .01 [$-$.08, .10] & .04 & +0.03 [$-$0.10, +0.16] & 0 & .02 & +0.01 [$-$0.12, +0.14] & 0 & $-$.03 & $-$0.04 [$-$0.17, +0.09] & 0 \\
NFC--Openness & .29 [.20, .37] & .36 & +0.08 [$-$0.04, +0.20] & P & .25 & $-$0.04 [$-$0.16, +0.09] & P & .25 & $-$0.03 [$-$0.16, +0.09] & P \\
Extraversion--Openness & .09 [$-$.00, .18] & .05 & $-$0.04 [$-$0.16, +0.09] & 0 & .08 & $-$0.01 [$-$0.14, +0.12] & 0 & .06 & $-$0.03 [$-$0.16, +0.10] & 0 \\
Agreeableness--Openness & .09 [.00, .18] & .13 & +0.03 [$-$0.09, +0.16] & P & .10 & +0.00 [$-$0.12, +0.13] & P & .11 & +0.02 [$-$0.11, +0.14] & P \\
Conscientiousness--Openness & .17 [.08, .26] & .22 & +0.05 [$-$0.08, +0.17] & P & .15 & $-$0.02 [$-$0.14, +0.11] & P & .13 & $-$0.04 [$-$0.16, +0.09] & P \\
CRT2--Neuroticism & $-$.03 [$-$.11, .06] & .07 & +0.10 [$-$0.03, +0.22] & 0 & .07 & +0.09 [$-$0.03, +0.22] & 0 & .06 & +0.08 [$-$0.05, +0.21] & 0 \\
bCRT--Neuroticism & $-$.02 [$-$.11, .07] & .02 & +0.04 [$-$0.08, +0.17] & 0 & .07 & +0.09 [$-$0.04, +0.22] & 0 & .07 & +0.09 [$-$0.04, +0.22] & 0 \\
NFC--Neuroticism & $-$.22 [$-$.31, $-$.13] & $-$.26 & $-$0.04 [$-$0.16, +0.09] & P & $-$.27 & $-$0.05 [$-$0.18, +0.07] & P & $-$.22 & +0.01 [$-$0.12, +0.13] & P \\
Extraversion--Neuroticism & $-$.33 [$-$.42, $-$.24] & $-$.33 & +0.01 [$-$0.12, +0.13] & P & $-$.32 & +0.02 [$-$0.10, +0.14] & P & $-$.28 & +0.06 [$-$0.06, +0.18] & P \\
Agreeableness--Neuroticism & $-$.32 [$-$.41, $-$.23] & $-$.40 & $-$0.07 [$-$0.19, +0.04] & P & $-$.44 & $-$0.12 [$-$0.23, +0.00] & P & $-$.36 & $-$0.04 [$-$0.16, +0.08] & P \\
Conscientiousness--Neuroticism & $-$.34 [$-$.42, $-$.25] & $-$.32 & +0.02 [$-$0.10, +0.14] & P & $-$.36 & $-$0.03 [$-$0.14, +0.09] & P & $-$.32 & +0.01 [$-$0.10, +0.13] & P \\
Neuroticism--Openness & $-$.00 [$-$.09, .09] & $-$.09 & $-$0.09 [$-$0.22, +0.04] & 0 & $-$.05 & $-$0.05 [$-$0.18, +0.08] & 0 & $-$.12 & $-$0.12 [$-$0.25, +0.01] & 0 \\
\bottomrule
\end{tabular}
\par\smallskip
\begin{minipage}{\linewidth}\footnotesize\raggedright
Spearman $\rho$; $\Delta\rho = \rho_{\mathrm{run}} - \rho_{\mathrm{human}}$; 95\% percentile bootstrap intervals, 10,000 resamples (seed 42), humans and agents resampled independently and the human frame resampled once for all runs; $N = 471$ per side. Verdict: P preserved, E exaggerated, W weakened, R reversed, I introduced, 0 no human target (the human interval includes 0), ? indeterminate; no correction across pairs (Section~\ref{subsec:statistical_analyses}).\par
\end{minipage}
\end{sidewaystable}

\begin{sidewaystable}
\centering
\scriptsize
\setlength{\tabcolsep}{3pt}
\caption{Trait-structure transfer, the four Claude Sonnet 4.6 conditions: per construct pair, the run's correlation, its difference from the human correlation, and the verdict.}
\label{tab:f4-b}
\begin{tabular}{@{}llrlcrlcrlcrlc@{}}
\toprule
 &  & \multicolumn{3}{c}{Claude Sonnet 4.6 · human · numeric} & \multicolumn{3}{c}{Claude Sonnet 4.6 · human · lexical} & \multicolumn{3}{c}{Claude Sonnet 4.6 · synthetic · numeric} & \multicolumn{3}{c}{Claude Sonnet 4.6 · synthetic · lexical} \\
\cmidrule(lr){3-5} \cmidrule(lr){6-8} \cmidrule(lr){9-11} \cmidrule(lr){12-14}
Pair & Human $\rho$ [95\% CI] & $\rho$ & $\Delta\rho$ [95\% CI] & V & $\rho$ & $\Delta\rho$ [95\% CI] & V & $\rho$ & $\Delta\rho$ [95\% CI] & V & $\rho$ & $\Delta\rho$ [95\% CI] & V \\
\midrule
NFC--Conscientiousness & .40 [.32, .48] & .46 & +0.06 [$-$0.05, +0.17] & P & .46 & +0.06 [$-$0.05, +0.17] & P & .44 & +0.04 [$-$0.07, +0.15] & P & .51 & +0.11 [+0.00, +0.21] & E \\
NFC--Extraversion & .13 [.03, .22] & .22 & +0.10 [$-$0.04, +0.24] & P & .20 & +0.07 [$-$0.06, +0.21] & P & .15 & +0.03 [$-$0.11, +0.16] & P & .17 & +0.04 [$-$0.09, +0.17] & P \\
NFC--Agreeableness & .08 [$-$.01, .18] & .15 & +0.07 [$-$0.07, +0.20] & 0 & .19 & +0.11 [$-$0.02, +0.25] & 0 & .25 & +0.17 [+0.04, +0.29] & I & .28 & +0.20 [+0.07, +0.32] & I \\
Extraversion--Agreeableness & .18 [.08, .27] & .19 & +0.01 [$-$0.12, +0.14] & P & .21 & +0.04 [$-$0.09, +0.17] & P & .21 & +0.04 [$-$0.09, +0.17] & P & .23 & +0.05 [$-$0.08, +0.18] & P \\
Extraversion--Conscientiousness & .18 [.09, .27] & .15 & $-$0.03 [$-$0.15, +0.10] & P & .13 & $-$0.04 [$-$0.17, +0.08] & P & .02 & $-$0.16 [$-$0.28, $-$0.03] & W & .08 & $-$0.10 [$-$0.23, +0.03] & ? \\
Agreeableness--Conscientiousness & .15 [.06, .24] & .11 & $-$0.04 [$-$0.18, +0.09] & P & .12 & $-$0.03 [$-$0.17, +0.10] & P & .11 & $-$0.04 [$-$0.18, +0.09] & P & .17 & +0.02 [$-$0.11, +0.15] & P \\
NFC--Openness & .29 [.20, .37] & .51 & +0.22 [+0.11, +0.34] & E & .57 & +0.29 [+0.17, +0.40] & E & .46 & +0.18 [+0.06, +0.30] & E & .55 & +0.26 [+0.15, +0.37] & E \\
Extraversion--Openness & .09 [$-$.00, .18] & .07 & $-$0.02 [$-$0.15, +0.12] & 0 & .08 & $-$0.01 [$-$0.14, +0.12] & 0 & .05 & $-$0.04 [$-$0.17, +0.09] & 0 & .09 & +0.00 [$-$0.13, +0.13] & 0 \\
Agreeableness--Openness & .09 [.00, .18] & .09 & $-$0.00 [$-$0.13, +0.13] & P & .12 & +0.02 [$-$0.10, +0.15] & P & .14 & +0.05 [$-$0.08, +0.17] & P & .14 & +0.04 [$-$0.09, +0.17] & P \\
Conscientiousness--Openness & .17 [.08, .26] & .12 & $-$0.05 [$-$0.18, +0.08] & P & .16 & $-$0.01 [$-$0.14, +0.12] & P & .07 & $-$0.10 [$-$0.23, +0.03] & ? & .16 & $-$0.01 [$-$0.14, +0.11] & P \\
NFC--Neuroticism & $-$.22 [$-$.31, $-$.13] & $-$.35 & $-$0.13 [$-$0.26, +0.00] & P & $-$.27 & $-$0.04 [$-$0.17, +0.09] & P & $-$.33 & $-$0.10 [$-$0.23, +0.03] & P & $-$.32 & $-$0.10 [$-$0.22, +0.03] & P \\
Extraversion--Neuroticism & $-$.33 [$-$.42, $-$.24] & $-$.37 & $-$0.04 [$-$0.17, +0.08] & P & $-$.31 & +0.03 [$-$0.09, +0.15] & P & $-$.33 & +0.00 [$-$0.12, +0.12] & P & $-$.28 & +0.05 [$-$0.07, +0.18] & P \\
Agreeableness--Neuroticism & $-$.32 [$-$.41, $-$.23] & $-$.35 & $-$0.03 [$-$0.15, +0.09] & P & $-$.36 & $-$0.04 [$-$0.17, +0.09] & P & $-$.41 & $-$0.09 [$-$0.21, +0.03] & P & $-$.40 & $-$0.08 [$-$0.19, +0.04] & P \\
Conscientiousness--Neuroticism & $-$.34 [$-$.42, $-$.25] & $-$.34 & $-$0.00 [$-$0.12, +0.12] & P & $-$.32 & +0.02 [$-$0.10, +0.14] & P & $-$.30 & +0.04 [$-$0.08, +0.16] & P & $-$.30 & +0.04 [$-$0.08, +0.16] & P \\
Neuroticism--Openness & $-$.00 [$-$.09, .09] & $-$.01 & $-$0.01 [$-$0.14, +0.13] & 0 & .01 & +0.01 [$-$0.12, +0.14] & 0 & $-$.05 & $-$0.05 [$-$0.18, +0.08] & 0 & $-$.07 & $-$0.07 [$-$0.19, +0.06] & 0 \\
\bottomrule
\end{tabular}
\par\smallskip
\begin{minipage}{\linewidth}\footnotesize\raggedright
Spearman $\rho$; $\Delta\rho = \rho_{\mathrm{run}} - \rho_{\mathrm{human}}$; 95\% percentile bootstrap intervals, 10,000 resamples (seed 42), humans and agents resampled independently and the human frame resampled once for all runs; $N = 471$ per side. Verdict: P preserved, E exaggerated, W weakened, R reversed, I introduced, 0 no human target (the human interval includes 0), ? indeterminate; no correction across pairs (Section~\ref{subsec:statistical_analyses}). The 13 pairs that involve CRT2 or bCRT are not estimable in these conditions (every agent scored 4/4 on both) and are omitted.\par
\end{minipage}
\end{sidewaystable}

\FloatBarrier
\subsection{BRT prevalence by run}
\label{app:brt_prevalence_by_run}
\begin{figure}[ht]
  \centering
  \includegraphics[width=\linewidth]{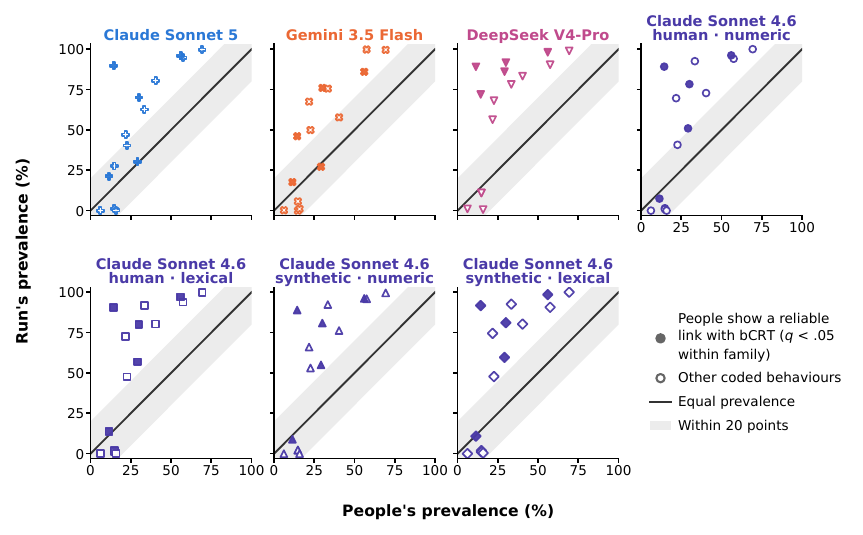}
  \caption{Prevalence of the 15 coded BRT behaviours in each run against people, one panel per run.}
  \label{fig:brt_prevalence_by_run}
\end{figure}

\begin{figure}[t]
  \centering
  \includegraphics[width=\linewidth]{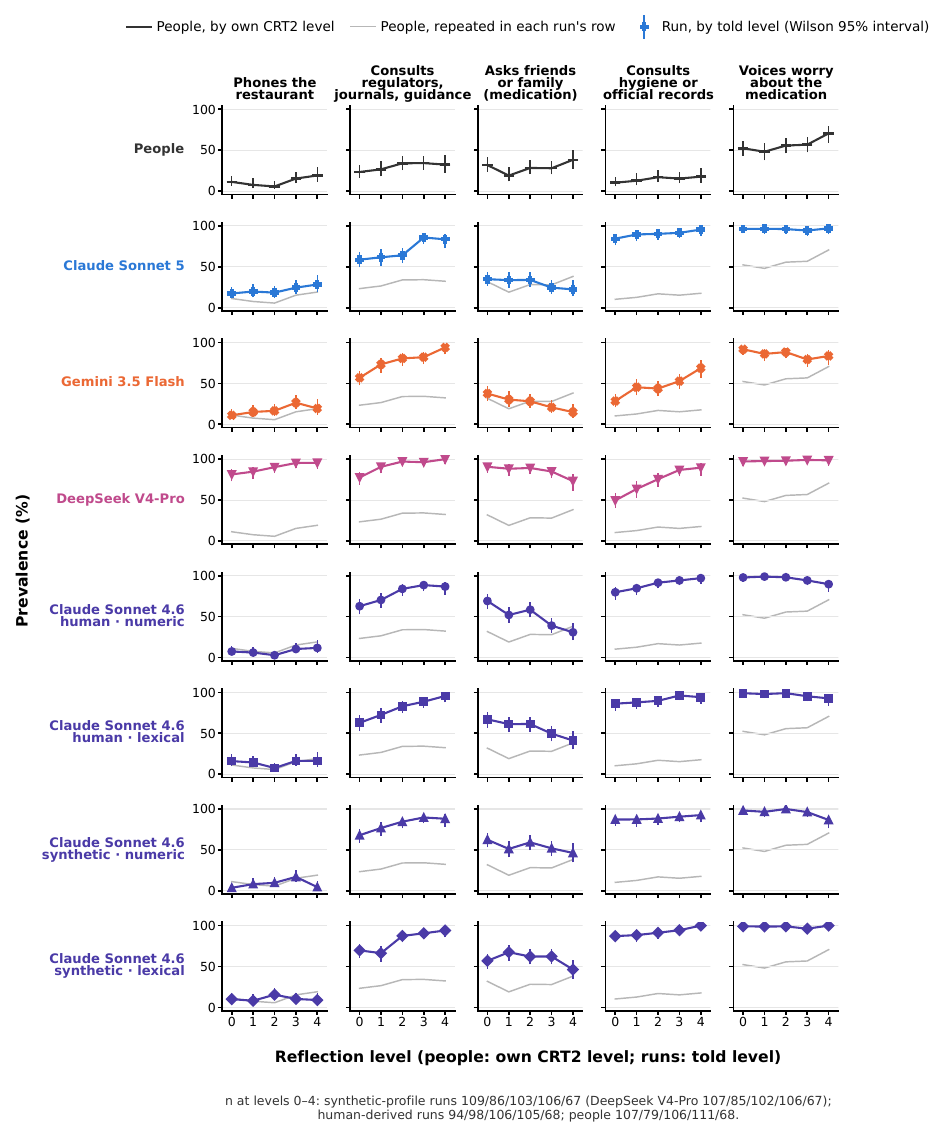}
  \caption{Does each behaviour change with the reflection level, in people and in each run? Prevalence of the five behaviours for which people show a reliable association with bCRT (Figure~\ref{fig:brt_behavioural_compression}B) at each reflection level 0--4; people's levels use the cutoffs the told levels were built on. $n$ per level is printed beneath the panels.}
  \label{fig:brt_level_prevalence}
\end{figure}

Figure~\ref{fig:brt_prevalence_by_run} repeats Figure~\ref{fig:brt_behavioural_compression}A for each run separately.
Figure~\ref{fig:brt_level_prevalence} gives the rates at each reflection level behind the steering results in Section~\ref{sec:results_brt_compression}.

\FloatBarrier
\subsection{BRT behaviour by run: prevalence, steering and transfer}
\label{app:brt_model_instability}

Tables~\ref{tab:brt-prevalence}, \ref{tab:brt-steering} and~\ref{tab:brt-transfer} give, for each of the 15 coded behaviours of Section~\ref{sec:results_brt_compression}, its prevalence in people and in each run, its steering in each run, and its transfer in the three runs whose bCRT scores vary. Unstable models (Appendix~\ref{app:formal_fidelity_definitions}) are fitted with Firth's penalised likelihood and reported, but carry no verdict and are left out of the tallies in Table~\ref{tab:scorecard}. Between 4 and 7 of the 15 behaviours are unstable in each run, because the runs show them in almost every response or almost none.

\begin{figure}[p]
  \centering
  \includegraphics[width=\linewidth]{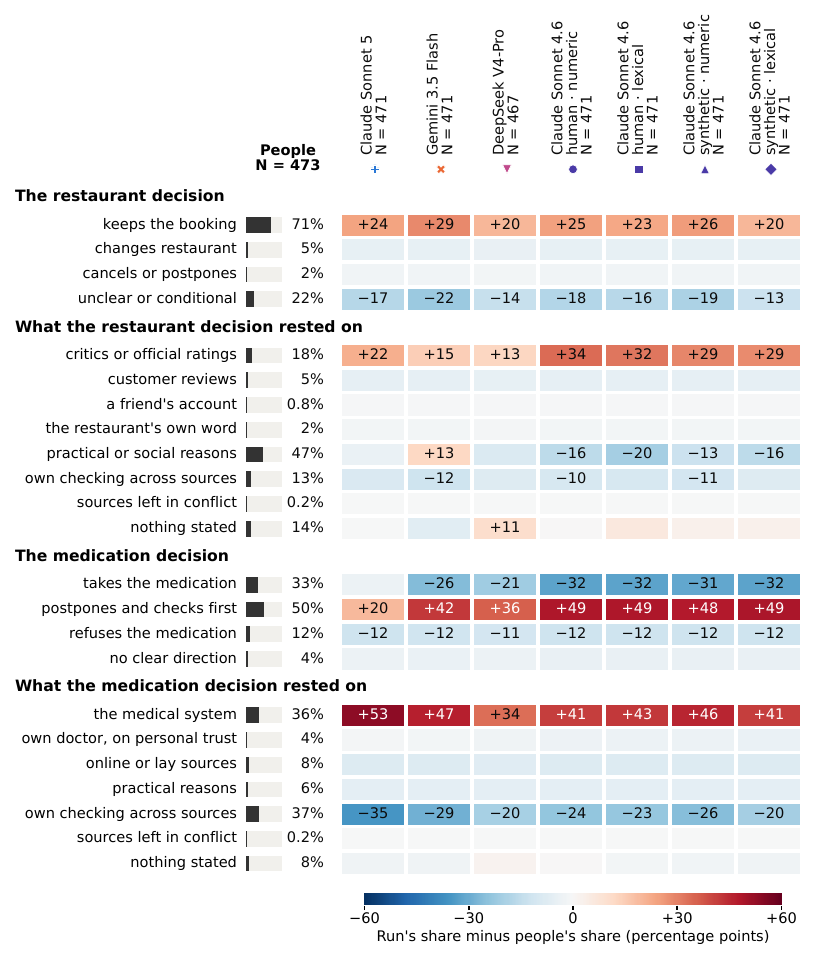}
  \caption{In which categories of the two decisions, and of what each decision rested on, do the runs differ most from people? The left column gives people's share of responses in each category, as a bar and a number. Each other cell gives a run's share minus people's, in percentage points: red where the run puts more responses in the category than people do, blue where it puts fewer, with the number printed where the gap is 10 points or more. Every group is coded with the same rubric; N is given with each column heading.}
  \label{fig:brt_categories}
\end{figure}

Figure~\ref{fig:brt_categories} shows where the runs' coded decisions depart from people's: every run puts more responses than people into keeping the booking, postponing the medication decision to check first, and resting that decision on the medical system, and fewer into resting it on their own checking across sources.

\begin{table}[t]
\centering
\footnotesize
\setlength{\tabcolsep}{3pt}
\caption{How often does each coded BRT behaviour occur in people and in each run? Prevalence (\%) of the 15 coded behaviours.}
\label{tab:brt-prevalence}
\begin{tabularx}{\linewidth}{@{}>{\hsize=1.00\hsize\raggedright\arraybackslash}Xrr@{}lr@{}lr@{}lr@{}lr@{}lr@{}lr@{}l@{}}
\toprule
 &  & \multicolumn{2}{c}{} & \multicolumn{2}{c}{} & \multicolumn{2}{c}{} & \multicolumn{8}{c}{Claude Sonnet 4.6} \\
\cmidrule(lr){9-16}
Behaviour & \begin{tabular}[b]{@{}c@{}}People \\ \% [95\% CI]\end{tabular} & \multicolumn{2}{c}{\begin{tabular}[b]{@{}c@{}}Claude \\ Sonnet 5\end{tabular}} & \multicolumn{2}{c}{\begin{tabular}[b]{@{}c@{}}Gemini \\ 3.5 Flash\end{tabular}} & \multicolumn{2}{c}{\begin{tabular}[b]{@{}c@{}}DeepSeek \\ V4-Pro\end{tabular}} & \multicolumn{2}{c}{\begin{tabular}[b]{@{}c@{}}human \\ numeric\end{tabular}} & \multicolumn{2}{c}{\begin{tabular}[b]{@{}c@{}}human \\ lexical\end{tabular}} & \multicolumn{2}{c}{\begin{tabular}[b]{@{}c@{}}synthetic \\ numeric\end{tabular}} & \multicolumn{2}{c}{\begin{tabular}[b]{@{}c@{}}synthetic \\ lexical\end{tabular}} \\
\midrule
\multicolumn{16}{@{}l}{\textit{Sources the respondent plans to consult}} \\
Phones the restaurant (the primary source) & 11.4 [8.9, 14.6] & 21.2 &  & 17.6 &  & 89.1 & \textsuperscript{C} & 7.4 &  & 13.8 &  & 8.9 &  & 10.8 &  \\
Consults regulators / journals / official guidance & 30.0 [26.1, 34.3] & 70.1 & \textsuperscript{C} & 76.0 & \textsuperscript{C} & 91.6 & \textsuperscript{C} & 78.3 & \textsuperscript{C} & 79.8 & \textsuperscript{C} & 80.9 & \textsuperscript{C} & 81.1 & \textsuperscript{C} \\
Asks friends or family (medication) & 29.2 [25.3, 33.4] & 30.4 &  & 27.2 &  & 86.1 & \textsuperscript{C} & 51.0 & \textsuperscript{C} & 56.9 & \textsuperscript{C} & 55.0 & \textsuperscript{C} & 59.7 & \textsuperscript{C} \\
Consults hygiene / official records & 14.4 [11.5, 17.8] & 89.8 & \textsuperscript{C} & 46.1 & \textsuperscript{C} & 71.9 & \textsuperscript{C} & 89.2 & \textsuperscript{C} & 90.7 & \textsuperscript{C} & 89.0 & \textsuperscript{C} & 91.7 & \textsuperscript{C} \\
Consults a qualified professional & 69.3 [65.0, 73.3] & 99.8 & \textsuperscript{CU} & 99.6 & \textsuperscript{CU} & 98.9 & \textsuperscript{CU} & 100.0 & \textsuperscript{CU} & 99.8 & \textsuperscript{CU} & 99.6 & \textsuperscript{CU} & 100.0 & \textsuperscript{CU} \\
Asks friends or family (restaurant) & 14.8 [11.9, 18.3] & 1.3 & \textsuperscript{U} & 0.0 & \textsuperscript{U} & 10.9 &  & 1.5 & \textsuperscript{U} & 2.1 & \textsuperscript{U} & 2.1 & \textsuperscript{U} & 1.9 & \textsuperscript{U} \\
Verifies the video's cited sources & 40.4 [36.1, 44.9] & 80.5 & \textsuperscript{C} & 57.7 &  & 83.3 & \textsuperscript{C} & 72.8 & \textsuperscript{C} & 80.3 & \textsuperscript{C} & 76.2 & \textsuperscript{C} & 80.3 & \textsuperscript{C} \\
\addlinespace[2pt]
\multicolumn{16}{@{}l}{\textit{Decision taken}} \\
Keeps the booking on evidence grounds & 57.5 [53.0, 61.9] & 94.7 & \textsuperscript{C} & 99.8 & \textsuperscript{CU} & 90.4 & \textsuperscript{C} & 94.1 & \textsuperscript{C} & 93.8 & \textsuperscript{C} & 96.0 & \textsuperscript{CU} & 90.7 & \textsuperscript{C} \\
Accepts on institutional grounds with a verification plan & 14.8 [11.9, 18.3] & 27.6 &  & 5.7 &  & 10.9 &  & 0.8 & \textsuperscript{U} & 1.3 & \textsuperscript{U} & 2.3 & \textsuperscript{U} & 1.1 & \textsuperscript{U} \\
Delays with a verification plan and a stated condition & 33.4 [29.3, 37.8] & 62.6 & \textsuperscript{C} & 75.6 & \textsuperscript{C} & 78.2 & \textsuperscript{C} & 92.6 & \textsuperscript{C} & 91.9 & \textsuperscript{C} & 92.4 & \textsuperscript{C} & 92.6 & \textsuperscript{C} \\
Acts on the claim with no doubt and no rule & 15.9 [12.8, 19.4] & 0.2 & \textsuperscript{U} & 1.1 & \textsuperscript{U} & 0.6 & \textsuperscript{U} & 0.0 & \textsuperscript{U} & 0.0 & \textsuperscript{U} & 0.0 & \textsuperscript{U} & 0.4 & \textsuperscript{U} \\
Refuses outright, worry-marked & 6.1 [4.3, 8.7] & 0.0 & \textsuperscript{U} & 0.2 & \textsuperscript{U} & 1.1 & \textsuperscript{U} & 0.0 & \textsuperscript{U} & 0.0 & \textsuperscript{U} & 0.0 & \textsuperscript{U} & 0.0 & \textsuperscript{U} \\
\addlinespace[2pt]
\multicolumn{16}{@{}l}{\textit{Emotion expressed}} \\
Voices worry about the medication & 56.0 [51.5, 60.4] & 96.0 & \textsuperscript{CU} & 86.0 & \textsuperscript{C} & 98.1 & \textsuperscript{CU} & 96.2 & \textsuperscript{CU} & 97.0 & \textsuperscript{CU} & 96.2 & \textsuperscript{CU} & 98.5 & \textsuperscript{CU} \\
Hostile rebuke of the unhelpful friend & 22.6 [19.1, 26.6] & 40.3 &  & 49.9 & \textsuperscript{C} & 68.1 & \textsuperscript{C} & 40.8 &  & 47.6 & \textsuperscript{C} & 53.1 & \textsuperscript{C} & 47.8 & \textsuperscript{C} \\
\addlinespace[2pt]
\multicolumn{16}{@{}l}{\textit{Appeals to fairness or loyalty}} \\
Appeals to fairness or loyalty toward the birthday friend or the restaurant & 21.8 [18.3, 25.7] & 47.1 & \textsuperscript{C} & 67.5 & \textsuperscript{C} & 56.3 & \textsuperscript{C} & 69.6 & \textsuperscript{C} & 72.6 & \textsuperscript{C} & 66.0 & \textsuperscript{C} & 74.5 & \textsuperscript{C} \\
\bottomrule
\end{tabularx}
\par\smallskip
\begin{minipage}{\linewidth}\footnotesize\raggedright
C: compressed, a gap of at least 20 percentage points with non-overlapping 95\% intervals; U: unstable, fewer than 20 events or non-events (Section~\ref{subsec:statistical_analyses}). N = 471 per run (DeepSeek V4-Pro 467); people N = 473.\par
\end{minipage}
\end{table}

\begin{sidewaystable}
\centering
\footnotesize
\setlength{\tabcolsep}{3pt}
\caption{Does each behaviour change with the reflection level the persona was told? Steering: odds ratio per told level, by run.}
\label{tab:brt-steering}
\begin{tabular}{@{}>{\raggedright\arraybackslash}p{1.5in}r@{}lr@{}lr@{}lr@{}lr@{}lr@{}lr@{}l@{}}
\toprule
 & \multicolumn{2}{c}{} & \multicolumn{2}{c}{} & \multicolumn{2}{c}{} & \multicolumn{8}{c}{Claude Sonnet 4.6} \\
\cmidrule(lr){8-15}
Behaviour & \multicolumn{2}{c}{\begin{tabular}[b]{@{}c@{}}Claude \\ Sonnet 5\end{tabular}} & \multicolumn{2}{c}{\begin{tabular}[b]{@{}c@{}}Gemini \\ 3.5 Flash\end{tabular}} & \multicolumn{2}{c}{\begin{tabular}[b]{@{}c@{}}DeepSeek \\ V4-Pro\end{tabular}} & \multicolumn{2}{c}{\begin{tabular}[b]{@{}c@{}}human \\ numeric\end{tabular}} & \multicolumn{2}{c}{\begin{tabular}[b]{@{}c@{}}human \\ lexical\end{tabular}} & \multicolumn{2}{c}{\begin{tabular}[b]{@{}c@{}}synthetic \\ numeric\end{tabular}} & \multicolumn{2}{c}{\begin{tabular}[b]{@{}c@{}}synthetic \\ lexical\end{tabular}} \\
\midrule
\multicolumn{15}{@{}l}{\textit{Sources the respondent plans to consult}} \\
Phones the restaurant (the primary source) & 1.17 [0.99, 1.37] &  & 1.25 [1.05, 1.49] &  & 1.56 [1.24, 1.98] &  & 1.20 [0.92, 1.55] &  & 1.01 [0.83, 1.23] &  & 1.23 [0.97, 1.55] &  & 1.01 [0.82, 1.25] &  \\
Consults regulators / journals / official guidance & 1.46 [1.25, 1.70] &  & 1.65 [1.39, 1.95] &  & 2.43 [1.73, 3.41] &  & 1.55 [1.30, 1.85] &  & 1.76 [1.45, 2.13] &  & 1.48 [1.23, 1.77] &  & 1.68 [1.39, 2.04] &  \\
Asks friends or family (medication) & 0.85 [0.74, 0.99] &  & 0.75 [0.65, 0.88] &  & 0.75 [0.61, 0.91] &  & 0.69 [0.60, 0.80] &  & 0.77 [0.67, 0.89] &  & 0.88 [0.77, 1.01] &  & 0.93 [0.81, 1.06] &  \\
Consults hygiene / official records & 1.31 [1.05, 1.65] &  & 1.43 [1.25, 1.65] &  & 1.80 [1.52, 2.13] &  & 1.66 [1.30, 2.13] &  & 1.37 [1.07, 1.75] &  & 1.15 [0.93, 1.42] &  & 1.53 [1.17, 1.99] &  \\
Consults a qualified professional & 2.16 [0.60, 108.32] & \textsuperscript{U} & 3.28 [0.94, 242.00] & \textsuperscript{U} & 7.47 [1.75, 804.11] & \textsuperscript{U} & 0.98 [0.02, 29.04] & \textsuperscript{U} & 1.38 [0.41, 8.00] & \textsuperscript{U} & 0.94 [0.35, 2.42] & \textsuperscript{U} & 0.98 [0.03, 22.39] & \textsuperscript{U} \\
Asks friends or family (restaurant) & 0.70 [0.35, 1.24] & \textsuperscript{U} & 1.02 [0.04, 39.49] & \textsuperscript{U} & 0.97 [0.79, 1.20] &  & 1.22 [0.71, 2.18] & \textsuperscript{U} & 0.85 [0.52, 1.34] & \textsuperscript{U} & 0.87 [0.55, 1.36] & \textsuperscript{U} & 1.01 [0.63, 1.63] & \textsuperscript{U} \\
Verifies the video's cited sources & 1.97 [1.61, 2.41] &  & 1.68 [1.45, 1.95] &  & 2.13 [1.69, 2.67] &  & 1.81 [1.52, 2.15] &  & 1.73 [1.43, 2.10] &  & 1.82 [1.52, 2.18] &  & 1.53 [1.28, 1.83] &  \\
\addlinespace[2pt]
\multicolumn{15}{@{}l}{\textit{Decision taken}} \\
Keeps the booking on evidence grounds & 0.93 [0.69, 1.24] &  & 0.66 [0.10, 2.16] & \textsuperscript{U} & 0.87 [0.69, 1.09] &  & 0.87 [0.65, 1.16] &  & 0.89 [0.67, 1.18] &  & 0.88 [0.63, 1.22] & \textsuperscript{U} & 0.63 [0.49, 0.80] &  \\
Accepts on institutional grounds with a verification plan & 0.98 [0.85, 1.14] &  & 0.91 [0.69, 1.21] &  & 0.97 [0.79, 1.20] &  & 0.62 [0.24, 1.27] & \textsuperscript{U} & 1.51 [0.84, 3.03] & \textsuperscript{U} & 1.02 [0.67, 1.58] & \textsuperscript{U} & 0.72 [0.34, 1.34] & \textsuperscript{U} \\
Delays with a verification plan and a stated condition & 1.11 [0.97, 1.27] &  & 1.03 [0.89, 1.20] &  & 1.15 [0.98, 1.35] &  & 1.14 [0.88, 1.48] &  & 1.24 [0.96, 1.60] &  & 0.92 [0.72, 1.19] &  & 1.11 [0.86, 1.43] &  \\
Acts on the claim with no doubt and no rule & 1.05 [0.30, 4.03] & \textsuperscript{U} & 0.48 [0.17, 0.97] & \textsuperscript{U} & 0.22 [0.00, 0.81] & \textsuperscript{U} & 1.02 [0.03, 40.61] & \textsuperscript{U} & 1.02 [0.03, 40.61] & \textsuperscript{U} & 1.02 [0.04, 39.49] & \textsuperscript{U} & 1.06 [0.41, 2.83] & \textsuperscript{U} \\
Refuses outright, worry-marked & 1.02 [0.04, 39.49] & \textsuperscript{U} & 0.46 [0.01, 1.66] & \textsuperscript{U} & 0.80 [0.40, 1.47] & \textsuperscript{U} & 1.02 [0.03, 40.61] & \textsuperscript{U} & 1.02 [0.03, 40.61] & \textsuperscript{U} & 1.02 [0.04, 39.49] & \textsuperscript{U} & 1.02 [0.04, 39.49] & \textsuperscript{U} \\
\addlinespace[2pt]
\multicolumn{15}{@{}l}{\textit{Emotion expressed}} \\
Voices worry about the medication & 0.96 [0.69, 1.33] & \textsuperscript{U} & 0.80 [0.66, 0.98] &  & 1.25 [0.78, 2.08] & \textsuperscript{U} & 0.57 [0.37, 0.83] & \textsuperscript{U} & 0.59 [0.37, 0.90] & \textsuperscript{U} & 0.56 [0.37, 0.82] & \textsuperscript{U} & 0.87 [0.50, 1.47] & \textsuperscript{U} \\
Hostile rebuke of the unhelpful friend & 1.01 [0.89, 1.16] &  & 0.90 [0.79, 1.03] &  & 0.78 [0.67, 0.90] &  & 0.85 [0.74, 0.97] &  & 0.85 [0.74, 0.98] &  & 1.02 [0.89, 1.16] &  & 0.87 [0.76, 0.99] &  \\
\addlinespace[2pt]
\multicolumn{15}{@{}l}{\textit{Appeals to fairness or loyalty}} \\
Appeals to fairness or loyalty toward the birthday friend or the restaurant & 0.87 [0.76, 1.00] &  & 0.89 [0.78, 1.03] &  & 0.98 [0.86, 1.12] &  & 1.08 [0.93, 1.25] &  & 1.08 [0.93, 1.25] &  & 0.99 [0.86, 1.13] &  & 0.95 [0.82, 1.11] &  \\
\bottomrule
\end{tabular}
\par\smallskip
\begin{minipage}{\linewidth}\footnotesize\raggedright
Logistic regression of each behaviour on the told level (0--4); Wald 95\% intervals, profile-likelihood intervals where the model is Firth-penalised (U rows). U: unstable (fewer than 20 events or non-events); its estimate is shown but carries no verdict. People have no told level; their per-level rates for the five behaviours with a reliable bCRT association are in Figure~\ref{fig:brt_level_prevalence}.\par
\end{minipage}
\end{sidewaystable}

\begin{table}[t]
\centering
\footnotesize
\setlength{\tabcolsep}{3pt}
\caption{Does people's link between bCRT and each behaviour carry over to the runs? Odds ratio per bCRT item in people and in the three runs whose bCRT scores vary, with the transfer verdict.}
\label{tab:brt-transfer}
\begin{tabularx}{\linewidth}{@{}>{\hsize=1.00\hsize\raggedright\arraybackslash}Xrrcrcrc@{}}
\toprule
 &  & \multicolumn{2}{c}{\begin{tabular}[b]{@{}c@{}}Claude \\ Sonnet 5\end{tabular}} & \multicolumn{2}{c}{\begin{tabular}[b]{@{}c@{}}Gemini \\ 3.5 Flash\end{tabular}} & \multicolumn{2}{c}{\begin{tabular}[b]{@{}c@{}}DeepSeek \\ V4-Pro\end{tabular}} \\
\cmidrule(lr){3-4} \cmidrule(lr){5-6} \cmidrule(lr){7-8}
Behaviour & \begin{tabular}[b]{@{}c@{}}People \\ OR [95\% CI]\end{tabular} & \begin{tabular}[b]{@{}c@{}}OR \\ {[95\% CI]}\end{tabular} & V & \begin{tabular}[b]{@{}c@{}}OR \\ {[95\% CI]}\end{tabular} & V & \begin{tabular}[b]{@{}c@{}}OR \\ {[95\% CI]}\end{tabular} & V \\
\midrule
\multicolumn{8}{@{}l}{\textit{Sources the respondent plans to consult}} \\
Phones the restaurant (the primary source) & 1.52 [1.21, 1.91] & 1.16 [1.01, 1.35] & T & 1.32 [1.04, 1.67] & T & 1.06 [0.86, 1.31] & F \\
Consults regulators / journals / official guidance & 1.33 [1.13, 1.57] & 1.40 [1.23, 1.59] & T & 1.54 [1.31, 1.80] & T & 1.23 [1.00, 1.51] & ? \\
Asks friends or family (medication) & 1.24 [1.05, 1.47] & 0.91 [0.80, 1.02] & F & 0.85 [0.72, 0.99] & F & 0.93 [0.75, 1.16] & F \\
Consults hygiene / official records & 1.28 [1.04, 1.57] & 1.24 [1.04, 1.49] & T & 1.39 [1.18, 1.63] & T & 1.34 [1.16, 1.55] & T \\
Consults a qualified professional & 1.16 [0.97, 1.37] & 1.22 [0.44, 3.60] & U & 2.61 [1.15, 23.33] & U & 1.22 [0.66, 1.91] & U \\
Asks friends or family (restaurant) & 0.98 [0.79, 1.21] & 0.76 [0.46, 1.22] & U & 0.63 [0.00, 3.81] & U & 0.96 [0.77, 1.18] & ? \\
Verifies the video's cited sources & 0.99 [0.85, 1.16] & 1.74 [1.49, 2.03] & I & 1.63 [1.39, 1.92] & I & 1.24 [1.06, 1.46] & I \\
\addlinespace[2pt]
\multicolumn{8}{@{}l}{\textit{Decision taken}} \\
Keeps the booking on evidence grounds & 1.00 [0.85, 1.16] & 0.97 [0.75, 1.26] & ? & 1.08 [0.18, 2.87] & U & 0.94 [0.73, 1.21] & ? \\
Accepts on institutional grounds with a verification plan & 1.09 [0.88, 1.34] & 0.95 [0.84, 1.08] & ? & 0.92 [0.68, 1.23] & ? & 0.89 [0.73, 1.08] & ? \\
Delays with a verification plan and a stated condition & 1.12 [0.95, 1.31] & 1.10 [0.98, 1.24] & ? & 1.01 [0.85, 1.20] & ? & 1.12 [0.96, 1.31] & ? \\
Acts on the claim with no doubt and no rule & 0.93 [0.75, 1.15] & 1.31 [0.50, 14.02] & U & 0.54 [0.29, 0.92] & U & 0.70 [0.41, 1.31] & U \\
Refuses outright, worry-marked & 0.88 [0.63, 1.22] & 0.84 [0.05, 6.44] & U & 0.44 [0.02, 1.18] & U & 0.82 [0.52, 1.51] & U \\
\addlinespace[2pt]
\multicolumn{8}{@{}l}{\textit{Emotion expressed}} \\
Voices worry about the medication & 1.44 [1.22, 1.70] & 0.96 [0.71, 1.28] & U & 0.70 [0.52, 0.93] & F & 0.70 [0.07, 1.26] & U \\
Hostile rebuke of the unhelpful friend & 1.10 [0.92, 1.31] & 0.96 [0.85, 1.08] & ? & 0.91 [0.78, 1.05] & ? & 0.92 [0.79, 1.07] & ? \\
\addlinespace[2pt]
\multicolumn{8}{@{}l}{\textit{Appeals to fairness or loyalty}} \\
Appeals to fairness or loyalty toward the birthday friend or the restaurant & 1.17 [0.97, 1.40] & 0.92 [0.82, 1.04] & F & 0.89 [0.76, 1.05] & F & 0.98 [0.85, 1.12] & ? \\
\bottomrule
\end{tabularx}
\par\smallskip
\begin{minipage}{\linewidth}\footnotesize\raggedright
Logistic regression on realised bCRT (per item); Wald 95\% intervals, profile-likelihood intervals for Firth-penalised (U) models. Verdicts test the run-minus-human slope difference (Section~\ref{subsec:statistical_analyses}): T transfers, F fails to transfer, I introduces (fails where people show no detectable association and the run's own interval excludes 1), ? indeterminate, U unstable. The four Claude Sonnet 4.6 conditions are not estimable: every agent scored 4/4 on the bCRT.\par
\end{minipage}
\end{table}

\FloatBarrier
\subsection{Two readings of the BRT pattern}
\label{app:brt_readings}

This section expands the reading of the BRT pattern in Section~\ref{sec:discussion}. One reading is that the runs express a stereotype of reflective behaviour rather than its graded human form. Postponing the medication decision until the claim is checked is by definition a reflective process, and every run over-produces it (Table~\ref{tab:brt-prevalence} and Figure~\ref{fig:brt_categories}). An alternative reading is an assistant default that recommends authoritative sources whatever the persona, which fits every run consulting a qualified professional in nearly every response (Figure~\ref{fig:brt_behavioural_compression}A; Table~\ref{tab:brt-prevalence}). However, the fact that the runs' behaviour changes across the reflection levels personas were told weighs against a persistent default: personas told the lowest level consult regulators in 57--78\% of responses, against 84--100\% at the highest, though both stay above people's 23--34\% at every level (Appendix Figure~\ref{fig:brt_level_prevalence}).

\FloatBarrier
\subsection{LIWC relationship fidelity}
\label{app:liwc_relationship_fidelity}

We evaluated whether the simulations reproduced human relationships between cognitive-reflection scores and theory-relevant LIWC markers in the BRT responses, using the seven focal categories of Appendix~\ref{app:liwc_analysis}; this backs the supplementary LIWC-transfer models of Section~\ref{subsec:statistical_analyses}.

In people, a higher bCRT score goes with longer responses, more present-focus and risk language, and fewer auxiliary verbs and less differentiation, each at $q < .05$ after correction within its family; its link to negative emotion does not survive correction ($q = .17$), and moral language shows none. On CRT2, only word count rises detectably.

\begin{figure}[p]
  \centering
  \includegraphics[width=\linewidth]{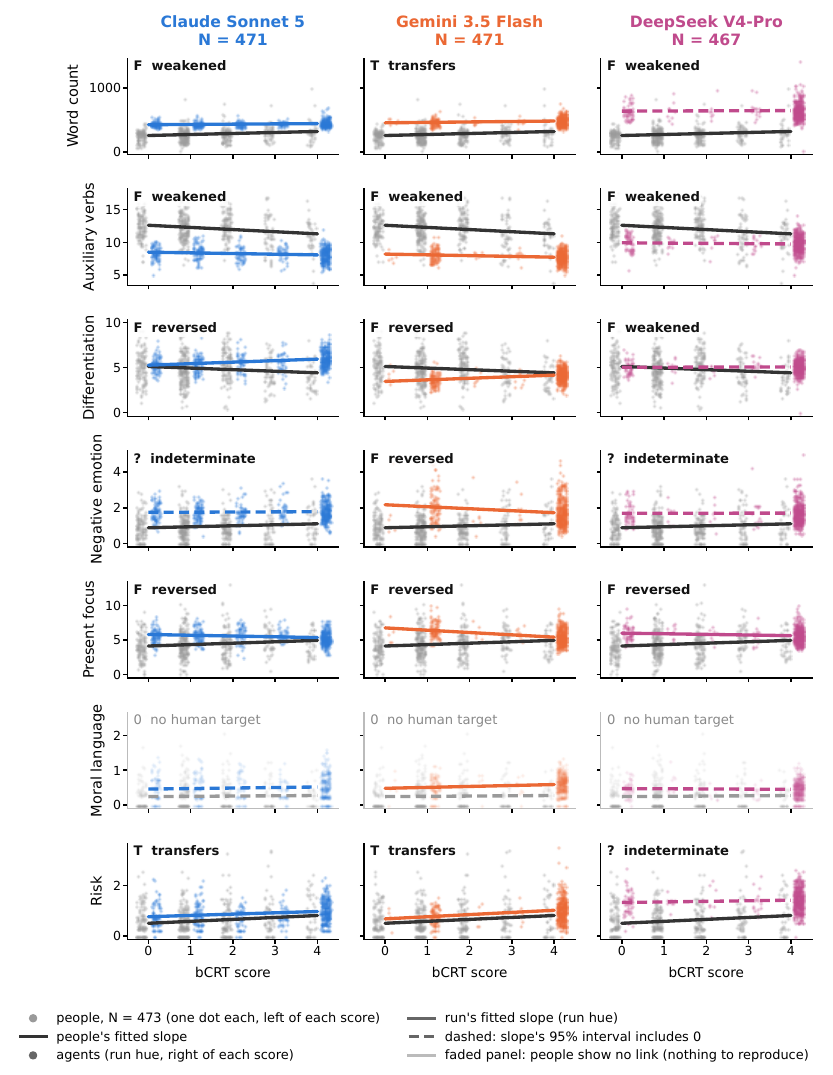}
  \caption{Does each run reproduce people's link between bCRT score and the language of their BRT responses? One column per run whose bCRT scores vary; one row per LIWC category of the theory-relevant set (Appendix~\ref{app:liwc_analysis}), chosen from prior work rather than for significance in people. Every person (grey) and agent (run hue) is a dot at its bCRT score, with people's fitted slope (black) and the run's (run hue), solid where the slope's 95\% interval excludes 0 and dashed where not. Each panel gives the verdict of the transfer rules (Appendix~\ref{app:formal_fidelity_definitions}): T transfers, F fails to transfer (weakened, reversed or exaggerated), ? indeterminate, I introduced (people show no link and the run differs), 0 no human target (faded: nothing to reproduce). People's link is judged by its uncorrected interval, as for the BRT. Rows share one scale; levels are shown, not tested. Tables~\ref{tab:liwc-slopes-1}--\ref{tab:liwc-slopes-3} give every slope and test. The four Claude Sonnet 4.6 runs are not shown: every agent scores 4 of 4.}
  \label{fig:liwc_dots_bcrt}
\end{figure}

\begin{figure}[p]
  \centering
  \includegraphics[width=\linewidth]{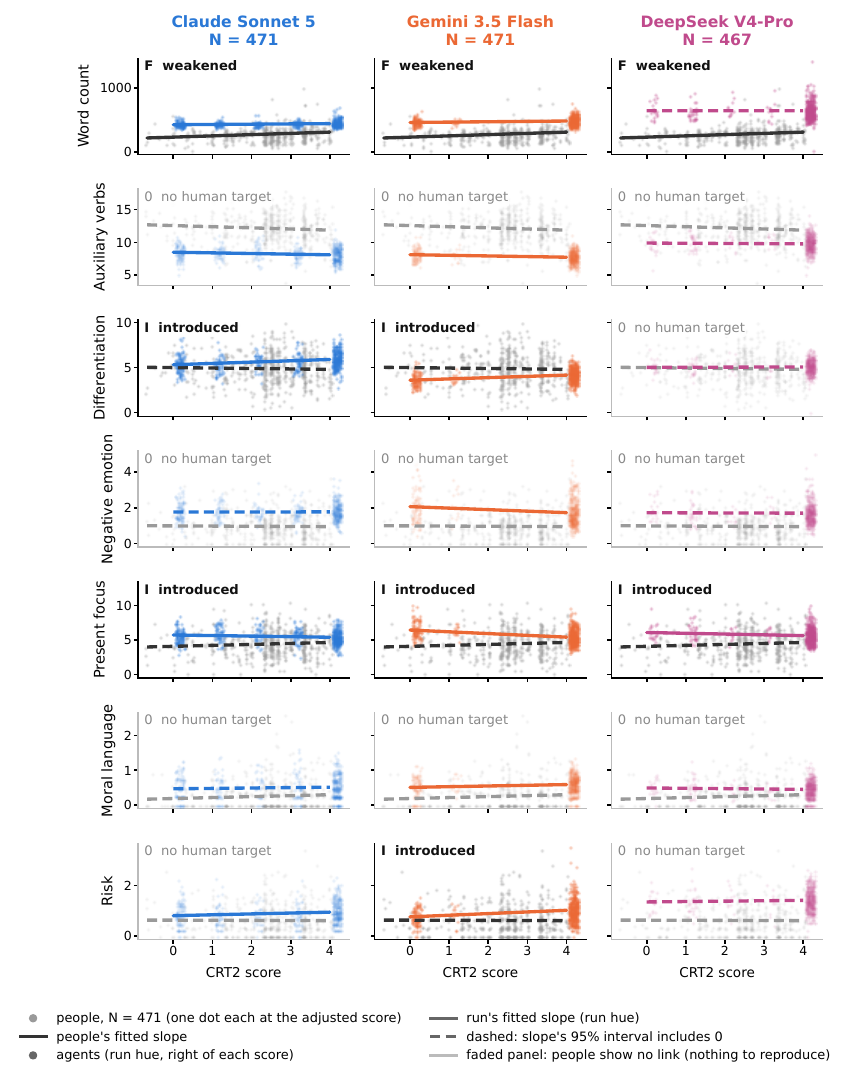}
  \caption{Does each run reproduce people's link between CRT2 score and the language of their BRT responses? As Figure~\ref{fig:liwc_dots_bcrt}, with people at their familiarity-adjusted CRT2 scores (Appendix~\ref{app:human_reference_data}), which are not whole numbers.}
  \label{fig:liwc_dots_crt2}
\end{figure}

The three runs whose CRT scores vary differ from people in 21 of their 42 comparisons at $p < .05$: 7 of 14 for Claude Sonnet 5, 8 for Gemini 3.5 Flash and 6 for DeepSeek V4-Pro (Tables~\ref{tab:liwc-slopes-1}--\ref{tab:liwc-slopes-3}). Judged by the transfer rules of Appendix~\ref{app:formal_fidelity_definitions}, of the 18 bCRT comparisons where people show a link by its uncorrected interval, 3 transfer, 12 fail to transfer (6 reversed, 6 weakened) and 3 are indeterminate; on CRT2, where people show a link only for word count, all three runs fail on it and introduce 6 links people do not show. Present-focus language falls as either test score rises in all three runs, where in people it rises with the bCRT, and differs from people in all six comparisons. On the bCRT, differentiation rises in Claude Sonnet 5 and Gemini 3.5 Flash and stays flat in DeepSeek V4-Pro, where in people it falls, and auxiliary verbs fall less steeply than in people; both differ in all three runs. Word-count slopes keep people's direction but are shallower, and differ in five of the six comparisons. No run differs from people on moral language, or on risk language with the bCRT. In the four Claude Sonnet 4.6 runs every agent scores 4 of 4 on both tests, so none of their 56 comparisons can be estimated, and their tables are not shown. Figures~\ref{fig:liwc_dots_bcrt} and~\ref{fig:liwc_dots_crt2} show every response behind these slopes: the agents' scores concentrate at full marks, and the runs' present-focus lines run against people's.

The three runs that vary on the two CRTs thus reproduce neither the size of people's links between reflection and language nor, for present focus and differentiation, their direction, although the same runs reproduce Need for Cognition and Big Five calibration (Section~\ref{sec:results_distribution_fidelity}).

\begin{table}[t]
\centering
\footnotesize
\setlength{\tabcolsep}{4pt}
\caption{LIWC relationship fidelity, Claude Sonnet 5.}
\label{tab:liwc-slopes-1}
\begin{tabularx}{\linewidth}{@{}l>{\hsize=1.00\hsize\raggedright\arraybackslash}Xrrrrrr@{}}
\toprule
\begin{tabular}[b]{@{}c@{}}Pre- \\ dictor\end{tabular} & LIWC outcome & \begin{tabular}[b]{@{}c@{}}Human \\ slope\end{tabular} & \begin{tabular}[b]{@{}c@{}}Agent \\ slope\end{tabular} & \begin{tabular}[b]{@{}c@{}}Differ- \\ ence\end{tabular} & 95\% CI & $p$ & \begin{tabular}[b]{@{}c@{}}Binned \\ RMSE\end{tabular} \\
\midrule
CRT2 & Word count & 19.634 & 4.203 & $-$15.431 & [$-$24.047, $-$6.815] & $<$.001 & 160.80 \\
CRT2 & Auxiliary verbs & $-$0.174 & $-$0.100 & 0.074 & [$-$0.111, 0.260] & .432 & 3.82 \\
CRT2 & Differentiation & $-$0.049 & 0.146 & 0.195 & [0.046, 0.344] & .010 & 0.99 \\
CRT2 & Negative emotion & $-$0.010 & 0.003 & 0.013 & [$-$0.053, 0.079] & .700 & 0.71 \\
CRT2 & Present focus & 0.140 & $-$0.081 & $-$0.221 & [$-$0.388, $-$0.055] & .009 & 1.43 \\
CRT2 & Moral language & 0.028 & 0.012 & $-$0.016 & [$-$0.055, 0.023] & .427 & 0.23 \\
CRT2 & Risk & $-$0.005 & 0.035 & 0.039 & [$-$0.019, 0.098] & .188 & 0.22 \\
bCRT & Word count & 15.797 & 4.319 & $-$11.478 & [$-$19.045, $-$3.912] & .003 & 140.01 \\
bCRT & Auxiliary verbs & $-$0.328 & $-$0.105 & 0.223 & [0.063, 0.384] & .007 & 3.52 \\
bCRT & Differentiation & $-$0.176 & 0.169 & 0.345 & [0.215, 0.474] & $<$.001 & 1.17 \\
bCRT & Negative emotion & 0.055 & 0.008 & $-$0.048 & [$-$0.106, 0.010] & .107 & 0.75 \\
bCRT & Present focus & 0.205 & $-$0.112 & $-$0.318 & [$-$0.463, $-$0.173] & $<$.001 & 1.10 \\
bCRT & Moral language & 0.008 & 0.015 & 0.007 & [$-$0.027, 0.041] & .676 & 0.22 \\
bCRT & Risk & 0.077 & 0.052 & $-$0.025 & [$-$0.076, 0.026] & .335 & 0.16 \\
\bottomrule
\end{tabularx}
\par\smallskip
\begin{minipage}{\linewidth}\footnotesize\raggedright
Slopes are from regressing the LIWC outcome on the predictor (CRT2 score or bCRT score); Difference = agent minus human, tested by the predictor$\times$source interaction. Binned RMSE compares binned agent and human means.\par
Archival identifiers: simulator run \texttt{CR\_\allowbreak MAIN\_\allowbreak SYNTH\_\allowbreak NUM\_\allowbreak SONNET5\_\allowbreak S42}; analysis report \texttt{cr13\_\allowbreak CR\_\allowbreak MAIN\_\allowbreak SYNTH\_\allowbreak NUM\_\allowbreak SONNET5\_\allowbreak S42}.\par
\end{minipage}
\end{table}

\begin{table}[t]
\centering
\footnotesize
\setlength{\tabcolsep}{4pt}
\caption{LIWC relationship fidelity, Gemini 3.5 Flash.}
\label{tab:liwc-slopes-2}
\begin{tabularx}{\linewidth}{@{}l>{\hsize=1.00\hsize\raggedright\arraybackslash}Xrrrrrr@{}}
\toprule
\begin{tabular}[b]{@{}c@{}}Pre- \\ dictor\end{tabular} & LIWC outcome & \begin{tabular}[b]{@{}c@{}}Human \\ slope\end{tabular} & \begin{tabular}[b]{@{}c@{}}Agent \\ slope\end{tabular} & \begin{tabular}[b]{@{}c@{}}Differ- \\ ence\end{tabular} & 95\% CI & $p$ & \begin{tabular}[b]{@{}c@{}}Binned \\ RMSE\end{tabular} \\
\midrule
CRT2 & Word count & 19.634 & 5.641 & $-$13.994 & [$-$23.167, $-$4.821] & .003 & 194.78 \\
CRT2 & Auxiliary verbs & $-$0.174 & $-$0.096 & 0.078 & [$-$0.103, 0.259] & .400 & 4.12 \\
CRT2 & Differentiation & $-$0.049 & 0.141 & 0.189 & [0.047, 0.332] & .009 & 1.15 \\
CRT2 & Negative emotion & $-$0.010 & $-$0.085 & $-$0.075 & [$-$0.154, 0.005] & .067 & 1.01 \\
CRT2 & Present focus & 0.140 & $-$0.259 & $-$0.398 & [$-$0.569, $-$0.228] & $<$.001 & 2.09 \\
CRT2 & Moral language & 0.028 & 0.020 & $-$0.007 & [$-$0.042, 0.028] & .676 & 0.34 \\
CRT2 & Risk & $-$0.005 & 0.065 & 0.069 & [0.012, 0.127] & .018 & 0.25 \\
bCRT & Word count & 15.797 & 7.033 & $-$8.764 & [$-$17.701, 0.172] & .055 & 181.51 \\
bCRT & Auxiliary verbs & $-$0.328 & $-$0.119 & 0.209 & [0.035, 0.384] & .019 & 4.03 \\
bCRT & Differentiation & $-$0.176 & 0.174 & 0.350 & [0.212, 0.488] & $<$.001 & 1.14 \\
bCRT & Negative emotion & 0.055 & $-$0.111 & $-$0.167 & [$-$0.244, $-$0.089] & $<$.001 & 0.85 \\
bCRT & Present focus & 0.205 & $-$0.337 & $-$0.543 & [$-$0.708, $-$0.378] & $<$.001 & 1.60 \\
bCRT & Moral language & 0.008 & 0.027 & 0.019 & [$-$0.015, 0.053] & .269 & 0.31 \\
bCRT & Risk & 0.077 & 0.085 & 0.008 & [$-$0.047, 0.064] & .773 & 0.20 \\
\bottomrule
\end{tabularx}
\par\smallskip
\begin{minipage}{\linewidth}\footnotesize\raggedright
Slopes are from regressing the LIWC outcome on the predictor (CRT2 score or bCRT score); Difference = agent minus human, tested by the predictor$\times$source interaction. Binned RMSE compares binned agent and human means.\par
Archival identifiers: simulator run \texttt{CR\_\allowbreak MAIN\_\allowbreak SYNTH\_\allowbreak NUM\_\allowbreak GEM35F\_\allowbreak MIN\_\allowbreak S42}; analysis report \texttt{cr13\_\allowbreak CR\_\allowbreak MAIN\_\allowbreak SYNTH\_\allowbreak NUM\_\allowbreak GEM35F\_\allowbreak MIN\_\allowbreak S42}.\par
\end{minipage}
\end{table}

\begin{table}[t]
\centering
\footnotesize
\setlength{\tabcolsep}{4pt}
\caption{LIWC relationship fidelity, DeepSeek V4-Pro.}
\label{tab:liwc-slopes-3}
\begin{tabularx}{\linewidth}{@{}l>{\hsize=1.00\hsize\raggedright\arraybackslash}Xrrrrrr@{}}
\toprule
\begin{tabular}[b]{@{}c@{}}Pre- \\ dictor\end{tabular} & LIWC outcome & \begin{tabular}[b]{@{}c@{}}Human \\ slope\end{tabular} & \begin{tabular}[b]{@{}c@{}}Agent \\ slope\end{tabular} & \begin{tabular}[b]{@{}c@{}}Differ- \\ ence\end{tabular} & 95\% CI & $p$ & \begin{tabular}[b]{@{}c@{}}Binned \\ RMSE\end{tabular} \\
\midrule
CRT2 & Word count & 19.634 & 0.045 & $-$19.589 & [$-$33.930, $-$5.248] & .007 & 378.91 \\
CRT2 & Auxiliary verbs & $-$0.174 & $-$0.031 & 0.143 & [$-$0.078, 0.364] & .206 & 2.09 \\
CRT2 & Differentiation & $-$0.049 & 0.011 & 0.060 & [$-$0.106, 0.227] & .477 & 0.35 \\
CRT2 & Negative emotion & $-$0.010 & $-$0.006 & 0.005 & [$-$0.076, 0.085] & .912 & 0.68 \\
CRT2 & Present focus & 0.140 & $-$0.113 & $-$0.253 & [$-$0.448, $-$0.058] & .011 & 1.45 \\
CRT2 & Moral language & 0.028 & $-$0.010 & $-$0.037 & [$-$0.077, 0.002] & .061 & 0.20 \\
CRT2 & Risk & $-$0.005 & 0.016 & 0.020 & [$-$0.047, 0.087] & .555 & 0.75 \\
bCRT & Word count & 15.797 & 1.997 & $-$13.801 & [$-$25.551, $-$2.050] & .021 & 355.59 \\
bCRT & Auxiliary verbs & $-$0.328 & $-$0.044 & 0.284 & [0.105, 0.463] & .002 & 2.03 \\
bCRT & Differentiation & $-$0.176 & 0.010 & 0.186 & [0.050, 0.322] & .007 & 0.78 \\
bCRT & Negative emotion & 0.055 & 0.004 & $-$0.052 & [$-$0.118, 0.014] & .124 & 0.66 \\
bCRT & Present focus & 0.205 & $-$0.092 & $-$0.297 & [$-$0.456, $-$0.138] & $<$.001 & 1.22 \\
bCRT & Moral language & 0.008 & $-$0.005 & $-$0.013 & [$-$0.045, 0.020] & .441 & 0.17 \\
bCRT & Risk & 0.077 & 0.023 & $-$0.054 & [$-$0.108, 0.001] & .054 & 0.70 \\
\bottomrule
\end{tabularx}
\par\smallskip
\begin{minipage}{\linewidth}\footnotesize\raggedright
Slopes are from regressing the LIWC outcome on the predictor (CRT2 score or bCRT score); Difference = agent minus human, tested by the predictor$\times$source interaction. Binned RMSE compares binned agent and human means.\par
DeepSeek V4-Pro BRT N = 467: four personas returned no usable open-ended response, and the provider has since retired the model snapshot.\par
Archival identifiers: simulator run \texttt{CR\_\allowbreak MAIN\_\allowbreak SYNTH\_\allowbreak NUM\_\allowbreak DSV4PRO\_\allowbreak TOG\_\allowbreak S42}; analysis report \texttt{cr13\_\allowbreak CR\_\allowbreak MAIN\_\allowbreak SYNTH\_\allowbreak NUM\_\allowbreak DSV4PRO\_\allowbreak TOG\_\allowbreak S42}.\par
\end{minipage}
\end{table}

\FloatBarrier

\clearpage
\section*{NeurIPS Paper Checklist}

\begin{enumerate}

\item {\bf Claims}
    \item[] Question: Do the main claims made in the abstract and introduction accurately reflect the paper's contributions and scope?
    \item[] Answer: \answerYes{}
    \item[] Justification: The abstract, Introduction, Results, and Discussion consistently frame the paper as an evaluation of cognitive-diversity fidelity in LLM persona simulations rather than as a psychometric validation paper. The main claims --- recovery of questionnaire-style trait structure alongside failures on the two CRTs, intuitive-lure fidelity, association structure, and downstream behavioural variation --- are directly supported by the reported distributional, calibration, association, and BRT analyses (Sections 1, 3, 4, 5, and 6).

\item {\bf Limitations}
    \item[] Question: Does the paper discuss the limitations of the work performed by the authors?
    \item[] Answer: \answerYes{}
    \item[] Justification: Section 6 explicitly discusses the main limitations of the study, including a full battery on four models, with further models and thinking regimes screened on the reasoning items only, the UK and US human reference sample, the stylised nature of the BRT scenarios, and dependence on an LLM-assisted coding rubric.

\item {\bf Theory assumptions and proofs}
    \item[] Question: For each theoretical result, does the paper provide the full set of assumptions and a complete (and correct) proof?
    \item[] Answer: \answerNA{}
    \item[] Justification: The paper does not present novel theoretical guarantees, proofs, or formal derivations. The fidelity targets in Section 3 and their formal notation in Appendix A.2 only define the evaluation setting and fidelity metrics.

\item {\bf Experimental result reproducibility}
    \item[] Question: Does the paper fully disclose all the information needed to reproduce the main experimental results of the paper to the extent that it affects the main claims and/or conclusions of the paper (regardless of whether the code and data are provided or not)?
    \item[] Answer: \answerYes{}
    \item[] Justification: The paper specifies the human reference sample, simulation conditions, prompting formats, models, task ordering, response-scoring pipeline, fidelity metrics, and statistical analyses in Sections 4 and Appendix A. The appendix additionally documents simulation conditions, synthetic-profile generation, scoring rules, BRT coding, and reproducibility procedures, including the release materials and configuration details.
    \item[] Guidelines:
    \begin{itemize}
        \item The answer \answerNA{} means that the paper does not include experiments.
        \item If the paper includes experiments, a \answerNo{} answer to this question will not be perceived well by the reviewers: Making the paper reproducible is important, regardless of whether the code and data are provided or not.
        \item If the contribution is a dataset and\slash or model, the authors should describe the steps taken to make their results reproducible or verifiable. 
        \item Depending on the contribution, reproducibility can be accomplished in various ways. For example, if the contribution is a novel architecture, describing the architecture fully might suffice, or if the contribution is a specific model and empirical evaluation, it may be necessary to either make it possible for others to replicate the model with the same dataset, or provide access to the model. In general. releasing code and data is often one good way to accomplish this, but reproducibility can also be provided via detailed instructions for how to replicate the results, access to a hosted model (e.g., in the case of a large language model), releasing of a model checkpoint, or other means that are appropriate to the research performed.
        \item While NeurIPS does not require releasing code, the conference does require all submissions to provide some reasonable avenue for reproducibility, which may depend on the nature of the contribution.
    \end{itemize}

\item {\bf Open access to data and code}
    \item[] Question: Does the paper provide open access to the data and code, with sufficient instructions to faithfully reproduce the main experimental results, as described in supplemental material?
    \item[] Answer: \answerYes{}
    \item[] Justification: Code is at \url{https://github.com/sion-w/CogSuite}, archived at \url{https://doi.org/10.5281/zenodo.24094423}, and data at \url{https://osf.io/h8d5s/}. The release includes de-identified schemas, configuration files, analysis scripts, fixed run metadata, and paper-asset generation scripts sufficient to reproduce the reported analyses from compatible inputs (Appendix A.10). Raw human free-text and individual-level participant data are not part of the public release; de-identified individual-level records are available from the authors (Appendix A.10).
    \item[] Guidelines:
    \begin{itemize}
        \item The answer \answerNA{} means that paper does not include experiments requiring code.
        \item Please see the NeurIPS code and data submission guidelines (\url{https://neurips.cc/public/guides/CodeSubmissionPolicy}) for more details.
    \end{itemize}

\item {\bf Experimental setting/details}
    \item[] Question: Does the paper specify all the training and test details (e.g., data splits, hyperparameters, how they were chosen, type of optimizer) necessary to understand the results?
    \item[] Answer: \answerYes{}
    \item[] Justification: Sections 4.1--4.4 and Appendix A specify the experimental setting in detail, including the human reference sample, synthetic-profile generation procedure, simulation conditions, prompt formats, models, task ordering, scoring rules, fidelity metrics, regression procedures, and interval methods. Appendix tables additionally document the exact simulation runs and configuration details.

\item {\bf Experiment statistical significance}
    \item[] Question: Does the paper report error bars suitably and correctly defined or other appropriate information about the statistical significance of the experiments?
    \item[] Answer: \answerYes{}
    \item[] Justification: The paper reports quantitative fidelity metrics, regression models, prevalence gaps, calibration statistics, and confidence-interval diagnostics throughout the Results and Appendix sections. Logistic-transfer analyses include instability and separation diagnostics where outcome compression invalidates conventional inference. Fidelity verdicts are read from 95\% intervals without correction across pairs or outcomes, and the paper states the expected rate of false departures (Section 4.4); Benjamini--Hochberg false-discovery-rate control within family identifies the human BRT and LIWC associations treated as reliable (Figure 4; Appendix B.8).
    \item[] Guidelines:
    \begin{itemize}
        \item The answer \answerNA{} means that the paper does not include experiments.
    \end{itemize}

\item {\bf Experiments compute resources}
    \item[] Question: For each experiment, does the paper provide sufficient information on the computer resources (type of compute workers, memory, time of execution) needed to reproduce the experiments?
    \item[] Answer: \answerNo{}
    \item[] Justification: The paper specifies the deployed models, simulation design, and run sizes, but does not provide detailed hardware, runtime, or total compute accounting because the experiments primarily used API-based inference rather than local model training. We instead provide the simulation configuration, run structure, and model-access details needed for conceptual reproducibility.
    \item[] Guidelines:
    \begin{itemize}
        \item The answer \answerNA{} means that the paper does not include experiments.
    \end{itemize}
    
\item {\bf Code of ethics}
    \item[] Question: Does the research conducted in the paper conform, in every respect, with the NeurIPS Code of Ethics \url{https://neurips.cc/public/EthicsGuidelines}?
    \item[] Answer: \answerYes{}
    \item[] Justification: The paper evaluates the limitations and failure modes of synthetic participants rather than advocating replacement of human participants. Human data collection followed institutional ethics procedures, and the release excludes identifiable participant information and raw free-text responses.
    \item[] Guidelines:
    \begin{itemize}
        \item The answer \answerNA{} means that the authors have not reviewed the NeurIPS Code of Ethics.
    \end{itemize}

\item {\bf Broader impacts}
    \item[] Question: Does the paper discuss both potential positive societal impacts and negative societal impacts of the work performed?
    \item[] Answer: \answerYes{}
    \item[] Justification: The Introduction and Discussion discuss both the potential utility of synthetic populations for behavioural and social-science research and the risks of over-trusting superficially plausible persona simulations that flatten meaningful cognitive and behavioural variation.
    \item[] Guidelines:
    \begin{itemize}
        \item The answer \answerNA{} means that there is no societal impact of the work performed.
    \end{itemize}
    
\item {\bf Safeguards}
    \item[] Question: Does the paper describe safeguards that have been put in place for responsible release of data or models that have a high risk for misuse (e.g., pre-trained language models, image generators, or scraped datasets)?
    \item[] Answer: \answerNA{}
    \item[] Justification: The paper does not release a new foundation model, scraped dataset, or generative system intended for deployment. The contribution is an evaluation framework and empirical audit of simulation fidelity.
    \item[] Guidelines:
    \begin{itemize}
        \item The answer \answerNA{} means that the paper poses no such risks.
    \end{itemize}

\item {\bf Licenses for existing assets}
    \item[] Question: Are the creators or original owners of assets (e.g., code, data, models), used in the paper, properly credited and are the license and terms of use explicitly mentioned and properly respected?
    \item[] Answer: \answerYes{}
    \item[] Justification: The paper cites and credits all major external assets used in the evaluation pipeline, including psychometric instruments, LIWC, prior datasets, and deployed model systems. Access to proprietary model APIs followed the corresponding provider usage terms.
    \item[] Guidelines:
    \begin{itemize}
        \item The answer \answerNA{} means that the paper does not use existing assets.
    \end{itemize}

\item {\bf New assets}
    \item[] Question: Are new assets introduced in the paper well documented and is the documentation provided alongside the assets?
    \item[] Answer: \answerYes{}
    \item[] Justification: The released materials document the schemas, scoring pipeline, configuration files, and analysis scripts used to generate the reported evaluation outputs. Reproducibility notes and simulation metadata are included in Appendix A.10.
    \item[] Guidelines:
    \begin{itemize}
        \item The answer \answerNA{} means that the paper does not release new assets.
    \end{itemize}

\item {\bf Crowdsourcing and research with human subjects}
    \item[] Question: For crowdsourcing experiments and research with human subjects, does the paper include the full text of instructions given to participants and screenshots, if applicable, as well as details about compensation (if any)? 
    \item[] Answer: \answerYes{}
    \item[] Justification: Section 4.1 and Appendix A.3 describe the human reference sample, its recruitment and eligibility, compensation (Prolific's recommended ethical hourly rate), consent and the task battery; the full participant instructions and materials are presented in the companion BRT study cited in Section 4.1.
    \item[] Guidelines:
    \begin{itemize}
        \item The answer \answerNA{} means that the paper does not involve crowdsourcing nor research with human subjects.
    \end{itemize}

\item {\bf Institutional review board (IRB) approvals or equivalent for research with human subjects}
    \item[] Question: Does the paper describe potential risks incurred by study participants, whether such risks were disclosed to the subjects, and whether Institutional Review Board (IRB) approvals (or an equivalent approval/review based on the requirements of your country or institution) were obtained?
    \item[] Answer: \answerYes{}
    \item[] Justification: The study was approved by the University of New South Wales Human Research Ethics Committee (approval reference iRECS7125), and every participant gave informed consent (Appendix A.3). The paper additionally documents privacy-related release constraints and excludes identifiable participant data from the released materials.
    \item[] Guidelines:
    \begin{itemize}
        \item The answer \answerNA{} means that the paper does not involve crowdsourcing nor research with human subjects.
    \end{itemize}

\item {\bf Declaration of LLM usage}
    \item[] Question: Does the paper describe the usage of LLMs if it is an important, original, or non-standard component of the core methods in this research? Note that if the LLM is used only for writing, editing, or formatting purposes and does \emph{not} impact the core methodology, scientific rigor, or originality of the research, declaration is not required.
    \item[] Answer: \answerYes{}
    \item[] Justification: LLM usage is central to the methodology and is described throughout Sections 4.1--4.3 and Appendix A, including the simulation protocol, persona prompting, response generation, and rubric-assisted scoring pipeline. The paper explicitly evaluates LLM-based synthetic participants as its primary research object.

\end{enumerate}

\end{document}